\documentclass{article}

    \PassOptionsToPackage{numbers, compress}{natbib}
\usepackage[preprint]{neurips_2026}

\usepackage[utf8]{inputenc} 
\usepackage[T1]{fontenc}    
\usepackage{hyperref}       
\usepackage{url}            
\usepackage{booktabs}       
\usepackage{amsfonts}       
\usepackage{nicefrac}       
\usepackage{microtype}      
\usepackage[dvipsnames]{xcolor}

\usepackage{amssymb}
\usepackage{amsmath}
\usepackage{graphicx}

\title{ActionMap: Robot Policy Learning via \\ Voxel Action Heatmap}

\author{%
  Pei Yang\textsuperscript{1,*} \quad
  Hai Ci\textsuperscript{1,*} \quad
  Yanzhe Chen\textsuperscript{1,*} \quad
  Qi Lv\textsuperscript{1} \quad
  Han Cai\textsuperscript{2} \quad
  Mike Zheng Shou\textsuperscript{1,\ddag} \\
  \textsuperscript{1}Show Lab, National University of Singapore \quad \quad \textsuperscript{2}NVIDIA
}

\begin{document}

\maketitle
\renewcommand{\thefootnote}{\fnsymbol{footnote}}
\footnotetext[1]{Equal contribution. \quad \quad \textsuperscript{\ddag}Corresponding author.}
\renewcommand{\thefootnote}{\arabic{footnote}}
\setcounter{footnote}{0}


\begin{figure}[htbp]
    \centering
    \includegraphics[width=0.95\linewidth]{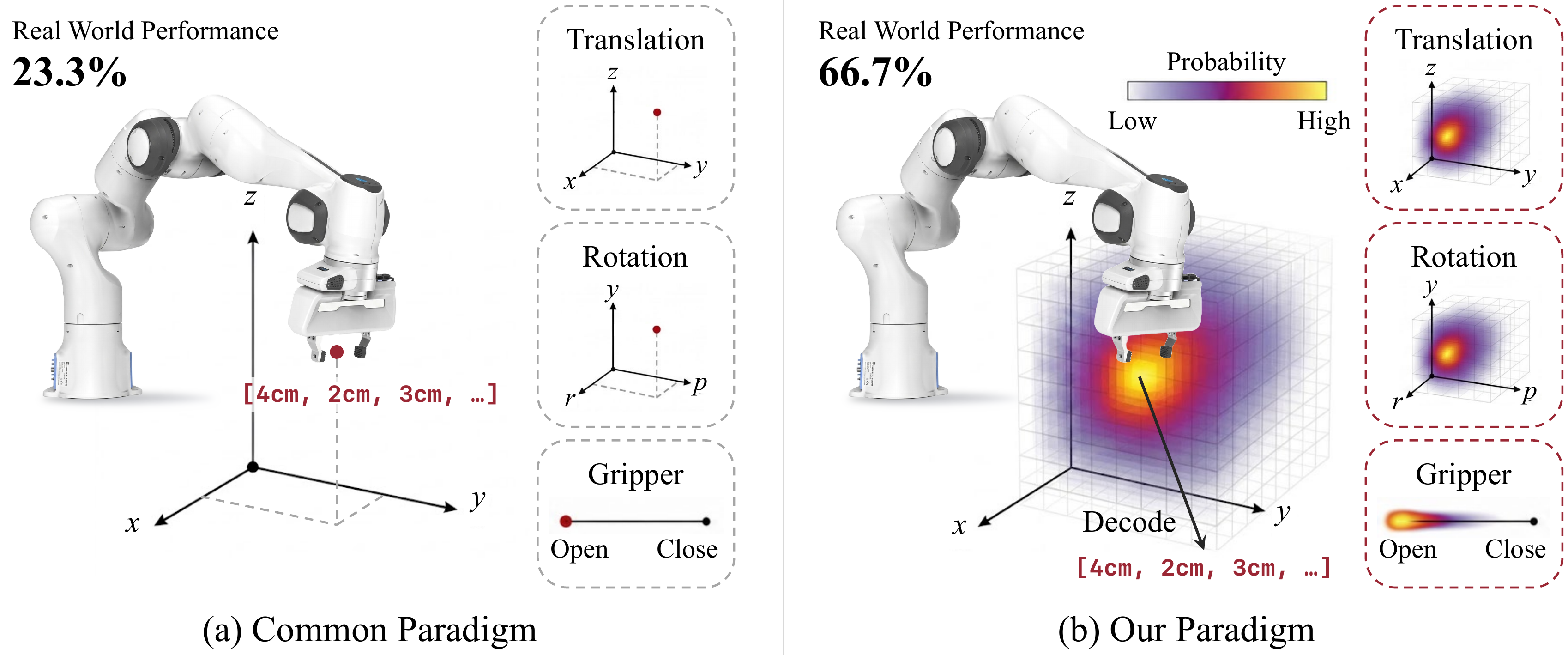}
    \caption{Overview of \textbf{ActionMap}. \textbf{(a)} The common paradigm directly predicts the next end-effector action as a single point in the continuous action space. \textbf{(b)} Our paradigm replaces this with a voxel heatmap of probabilities over the action space, and recovers the next action by decoding from this distribution. The heatmap is visualized at the end-effector pose for clarity.}
    \label{fig:teaser}
\end{figure}

\begin{abstract}
    Vision-language-action (VLA) models have advanced rapidly across backbones, training recipes, and data scale, yet the action decoder, which converts the backbone's hidden state into a continuous control signal, has barely changed and remains a single-point predictor across the majority of current VLAs. Whether implemented via autoregressive token bins, L1 regression, or flow-matching denoising, the resulting decoder treats the action space as unstructured, leaving the geometric proximity of neighboring actions unexploited during training. To advance this, we introduce \textbf{ActionMap}, a voxel heatmap action head that drops into an existing VLA in place of its native action decoder. For each new action, the head predicts a voxel heatmap over the action space, where each voxel directly stores the probability of the corresponding action. Across LIBERO simulation and real-world Franka manipulation, our heatmap head surpasses two architecturally distinct backbones at matched training steps (e.g., $+8.2\%$ over OpenVLA-OFT's L1 regression head on the LIBERO four-suite average), converges at comparable or faster rates on both backbones, and remains markedly more data-efficient at low training data. The cross-backbone consistency indicates that action representation is a real lever for VLA performance, distinct from further backbone or recipe scaling. Project Page: \url{https://showlab.github.io/ActionMap/}.
\end{abstract}


\section{Introduction}
\label{sec:intro}

Vision-language-action (VLA) models have rapidly become the dominant approach to instruction-following robotic manipulation, scaling from autoregressive token-based control \cite{rt1, brohan2023rt2, OpenVLA} to parallel-decoding action experts \cite{OpenVLA-OFT} and continuous denoising heads \cite{black2024pi0, Pi-0.5, gr00tn1_2025}. Across this rapid progress on backbones, training recipes, and data scale, the action decoder itself, the component that converts the backbone's hidden state into a control signal, has barely changed and remains a single-point predictor across nearly all current systems. The result is no spatial structure for the network to exploit during training, no soft probabilistic supervision over neighboring actions, and no differentiable handle on multimodal demonstrations, all of which can limit data efficiency, convergence, and end-effector precision in ways that further backbone scaling alone is unlikely to fully address.

Outside the VLA setting, action heads with explicit spatial outputs, such as voxelized one-hot heads \cite{shridhar2022perceiver, james2022coarse} and multi-view 2D heatmaps \cite{goyal2023rvt, goyal2024rvt2}, have consistently outperformed direct regression on manipulation benchmarks. Yet no prior work has integrated such a head into a large pretrained VLA, leaving this established gain on the table. Our key observation is that the end-effector action space has rich geometric structure, with neighboring actions carrying similar physical meaning, and that a soft distributional output naturally captures this structure: by predicting a probability distribution over a discretized voxel heatmap grid for each action component, the model can explicitly assign mass to neighboring action values, soft-supervise via Gaussian-blob targets, and recover continuous actions through differentiable soft-argmax decoding.

Building on this observation, we introduce \textbf{ActionMap} (Fig.~\ref{fig:teaser}), a voxel heatmap action head that drops into an VLA in place of its native action decoder. For each new action, the head predicts three voxel heatmaps, one over a 3D translation grid, one over a 3D rotation grid, and one over the gripper, where each voxel directly stores the probability of the corresponding action. We train these heatmaps against a softmax-normalized Gaussian blob centered on the ground-truth action via cross-entropy. At inference, a continuous action can be recovered from each heatmap by hard argmax (as in image classification), top-$k$ soft argmax, or any other distribution-decoding rule. Our heatmap head directly replaces the output action decoder of an existing VLA, whether OpenVLA-OFT's L1 regression head or $\pi_{0.5}$'s flow-matching action expert, and we validate it on both backbones in LIBERO simulation as well as on real-world Franka manipulation.

Across experiments, our heatmap head delivers four advantages over the corresponding baselines. \textbf{Better performance:} on the LIBERO four-suite average, our heatmap head improves over OpenVLA-OFT's L1 regression head by +8.2\% (97.3\% vs.\ 89.1\%) and over $\pi_{0.5}$'s flow-matching head by +1.6\% (98.5\% vs.\ 96.9\%) at matched training steps; on real-world Franka manipulation, our heatmap head wins on every task at full data (20/30 vs.\ 7/30 trials over OpenVLA-OFT's L1 regression head) and reduces end-effector grasp position error by 2 to 3 times. \textbf{Stronger data efficiency:} at 10\% LIBERO-Spatial demonstrations, our heatmap head holds 93.2\% while OpenVLA-OFT's L1 regression head collapses to 67.2\%. \textbf{Faster convergence:} across two backbones and two data regimes, our heatmap head consistently converges faster than the baseline, except for one setting where convergence speed is comparable. \textbf{Backbone generality:} our heatmap head applies to two mainstream VLA backbones, OpenVLA-OFT (L1 regression) and $\pi_{0.5}$ (flow matching), without changing any other network component, indicating that the design is not tied to a particular action paradigm.

Our contributions are summarized as follows:

\begin{itemize}
    \item \textbf{Voxel heatmap action head.} A drop-in replacement for the output action decoder of any VLA, trained via cross-entropy against a softmax-normalized Gaussian-blob target over translation, rotation, and gripper voxel grids, and decoded at inference by hard argmax, top-$k$ soft argmax, or any other logit-decoding rule.
    \item \textbf{Simulation validation across two architectures.} On LIBERO, our heatmap head replaces OpenVLA-OFT's L1 regression head and $\pi_{0.5}$'s flow-matching action expert without changing any other component, improving the four-suite average by +8.2\% over OpenVLA-OFT and +1.6\% over $\pi_{0.5}$ at matched training steps, training faster on both backbones, and retaining a +26.0\% gap over OpenVLA-OFT's L1 regression head at 10\% training data.
    \item \textbf{Real-world validation.} On three manipulation tasks (Pick, Sweep, Insert) on a Franka Research 3 robot, our heatmap head outperforms OpenVLA-OFT's L1 regression head on every task at full data (pooled 20/30 vs.\ 7/30 trials), retains its win on Pick and Sweep at 50 demonstrations, and reduces end-effector grasp position error by 2 to 3 times.
\end{itemize}

\section{Related Work}
\label{sec:related_works}

\paragraph{Robotic Manipulation and VLAs.} Vision-language-action (VLA) models combine pretrained vision-language backbones with imitation learning to produce instruction-following manipulation policies \cite{rt1, brohan2023rt2, OpenVLA, OpenVLA-OFT, octo, black2024pi0, Pi-0.5, gr00tn1_2025, roboflamingo, hpt, rdt, tinyvla, zheng2025x}, building on continuous behavior-cloning architectures such as diffusion and chunked-transformer policies \cite{chi2023diffusion, zhao2023act}, trained on increasingly diverse robot datasets \cite{padalkar2023oxe, droid, bridgev2}, and evaluated on standardized manipulation benchmarks \cite{LIBERO, robocasa, calvin}. By comparison, the action representation itself, namely how the backbone's hidden states are converted into a continuous control signal, has received little dedicated study and remains a single-point output across nearly all current systems.

\paragraph{VLA Action Representation.} Action representations in current VLAs fall into two narrow families. Discrete decoders bin each action dimension independently into uniform tokens and recast control as language modeling \cite{brohan2023rt2, OpenVLA}, which inherits mature autoregressive infrastructure but introduces hard quantization and per-dimension independence. Continuous decoders instead regress the action vector via $\ell_1$ loss \cite{OpenVLA-OFT} or draw a sample from a denoising process such as diffusion or flow matching \cite{chi2023diffusion, black2024pi0, Pi-0.5, gr00tn1_2025}, which removes quantization but produces a single unstructured point estimate per inference and offers no explicit spatial supervision over the action space.

In response to this lack of spatial structure, action heads with explicit spatial outputs have been explored outside the VLA setting. PerAct \cite{shridhar2022perceiver} voxelizes the workspace and predicts the next end-effector pose as a one-hot voxel; C2F-ARM \cite{james2022coarse} refines this hierarchically with a coarse-to-fine voxel grid; RVT \cite{goyal2023rvt} and RVT-2 \cite{goyal2024rvt2} re-render the workspace into virtual orthographic views and predict 2-D heatmaps per view. These heads consistently outperform direct regression on their target benchmarks, but they are tightly coupled to small specialized backbones such as the Perceiver-IO transformer or multi-view CNNs, and have not been transferred to large pretrained VLAs. A separate line of work introduces 3-D structure on the observation side rather than the action side \cite{huang2023voxposer, li2025bridgevla}. Across these directions, no prior work has explored a spatially structured 3-D voxel action representation on a large pretrained VLA.

\paragraph{Heatmap-Based Spatial Prediction in Pose Estimation.} The same shift from point regression to spatial heatmap prediction has played out before in human pose estimation. Early work directly regressed joint coordinates with a CNN \cite{toshev2014deeppose}, but accuracy plateaued and the field rapidly converged on predicting per-joint Gaussian-blob heatmaps \cite{wei2016convolutional, newell2016stacked, xiao2018simple, sun2019hrnet}, which became the dominant representation across both 2-D~\cite{wei2016convolutional,newell2016stacked,xiao2018simple, sun2019hrnet} and 3-D~\cite{ma2021context} pose estimation. Subsequent work bridged the heatmap-versus-regression divide by recovering continuous coordinates from heatmaps via differentiable soft-argmax \cite{sun2018integral}, achieving sub-pixel precision while retaining the dense gradient signal of heatmap supervision. Inspired by this trajectory, our heatmap action head adopts the same design for VLA action prediction: a Gaussian-blob target over a voxel heatmap grid replaces the point-estimate regression, and a top-$k$ soft-argmax recovers a continuous action at inference.

\section{Methodology}

\begin{figure}
    \centering
    \includegraphics[width=\linewidth]{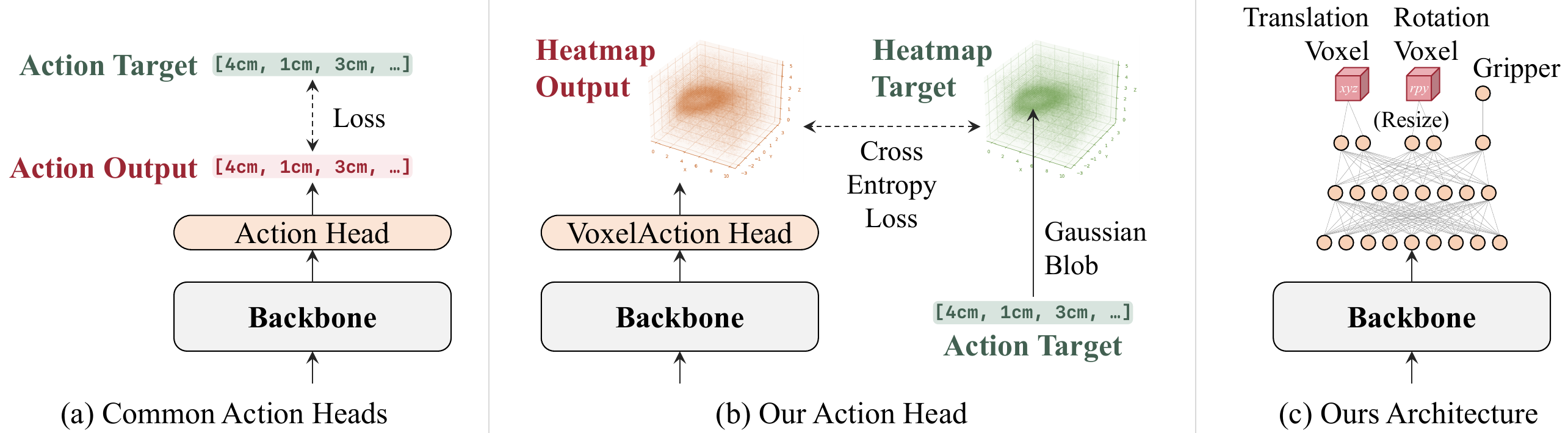}
    \caption{Training architecture of our voxel heatmap action head. \textbf{(a)} A common action head produces a single point estimate of the action and is trained directly against the ground-truth action vector. \textbf{(b)} Our voxel heatmap head instead generates a probability map over the action space and is trained against a Gaussian-blob target derived from the ground-truth action via cross-entropy. \textbf{(c)} Concretely, the head is a pure-MLP architecture with residual connections; a shared trunk branches into three parallel heads that produce a translation voxel heatmap, a rotation voxel heatmap, and a gripper open/close probability, with the translation and rotation flat outputs reshaped into 3D voxel grids.}
    \label{fig:methodology_model}
\end{figure}

\subsection{Problem Formulation}
\label{sec:methodology_problem}

In a VLA policy, the action decoder produces the next robot action $\mathbf{a}$ from the backbone's hidden state. We aim to make the next-action distribution explicit and spatially grounded at the output. This approach contrasts with mainstream decoders, which are either point estimates (regression~\cite{OpenVLA-OFT}, autoregressive tokens~\cite{OpenVLA}) or implicit-denoising endpoints (flow matching~\cite{Pi-0.5}). 

Formally, given an observation $\mathbf{o}$ and a language instruction $\ell$, the action head explicitly outputs the distribution $\Pr(\mathbf{a} \mid \mathbf{o}, \ell)$ \textbf{at every location within a finite action range}. The next action $\mathbf{a}$ is usually seven-dimensional (three translation, three rotation, one gripper). Representing this full 7-D joint as a single voxel heatmap grid is intractable, so we factor it into a 3-D translation distribution $\Pr(x, y, z)$, a 3-D rotation distribution $\Pr(\phi, \theta, \psi)$ over Euler angles, and a binary gripper distribution $\Pr(g)$. The spatial components $(x, y, z, \phi, \theta, \psi)$ are parameterized as either absolute end-effector coordinates or per-step velocities $(v_x, v_y, v_z, v_\phi, v_\theta, v_\psi)$ (i.e., deltas).

\subsection{Heatmap Action Head Architecture}
\label{sec:methodology_architecture}

We realize these factored distributions using a \textbf{voxel heatmap action head} that drops into an existing VLA in place of its native action decoder. The rest of the policy is preserved. This head outputs each of the three component distributions as a discretized voxel heatmap grid (Fig.~\ref{fig:methodology_model}c). Crucially, each voxel in the translation and rotation grids (and each of the two bins in the gripper grid) directly contains the probability mass for the action taking that specific value, such as $\Pr(x, y, z)$ for a translation voxel and $\Pr(\phi, \theta, \psi)$ for a rotation voxel. 

To generate these grids for a single action step, the head reads the backbone's last-layer hidden states at the corresponding action-token positions. It feeds these states, in parallel, to three independent MLPs that produce logits over the translation grid (shape $N_x{\times}N_y{\times}N_z$), the rotation grid (shape $N_\phi{\times}N_\theta{\times}N_\psi$), and the two gripper bins. Following recent VLAs~\cite{OpenVLA-OFT, Pi-0.5}, the grids in our experiments are laid out over the delta-action space (per-step velocity). Finally, modern VLAs typically return a sequence of $T$ actions per forward pass. To support this chunked output, the head reads from all $T$ action-token positions simultaneously and produces $T$ such triplets of grids in one shot.

\subsection{Training with Soft-Label Supervision}
\label{sec:methodology_training}

At training time, we convert each ground-truth action component into a Gaussian blob over the corresponding voxel heatmap grid (Fig.~\ref{fig:methodology_model}b). This blob is a softmax-normalized Gaussian centered on the ground-truth action and falling off with distance. We then train the head's output to match this soft target via cross-entropy. For a ground-truth action component $\mathbf{a}^*_c$ on branch $c$, with grid $\mathcal{B}_c$ and bin centers $b$, the target distribution (Gaussian blob) is given by
\begin{equation}
    \underbrace{q_\sigma(b;\,\mathbf{a}^*_c)}_{\text{target}}
    \;=\;
    \frac{\exp\!\left(-\|b - \mathbf{a}^*_c\|^2 / 2\sigma^2\right)}
         {\sum_{b' \in \mathcal{B}_c} \exp\!\left(-\|b' - \mathbf{a}^*_c\|^2 / 2\sigma^2\right)}.
    \label{eq:gaussian_target}
\end{equation}

\noindent The head is trained by minimizing the soft-label cross-entropy between this target and its predicted softmax distribution. This loss is summed across the three branches:
\begin{equation}
    \mathcal{L}
    \;=\;
    -\sum_{c \in \{\mathrm{trans},\,\mathrm{rot},\,\mathrm{grip}\}}
    \sum_{b \in \mathcal{B}_c}
        \underbrace{q_\sigma(b;\,\mathbf{a}^*_c)}_{\text{target}}
        \log \underbrace{p_\theta^{(c)}(b \mid \mathbf{h})}_{\text{prediction}},
    \label{eq:loss}
\end{equation}

\noindent where $p_\theta^{(c)}(\cdot \mid \mathbf{h})$ is the head's predicted softmax on branch $c$ given the action-token hidden state $\mathbf{h}$. The gripper branch is a degenerate case in which $|\mathcal{B}_{\mathrm{grip}}|{=}2$, and its target collapses to a one-hot vector at the ground-truth label. The per-slot loss is summed over all $T$ chunk slots in the VLA forward pass. Ultimately, this formulation replaces coordinate regression with heatmap-based classification, a strategy widely observed to optimize more cleanly in spatial prediction tasks~\cite{wei2016convolutional, newell2016stacked, sun2019hrnet, sun2018integral}.

\subsection{Inference via Top-$k$ Decoding}
\label{sec:methodology_inference}

At inference, we must pick a single action from each predicted distribution. We call this step \emph{decoding}. For an action head's output logits $z$ over the voxel heatmap grid, a hard argmax selects the voxel center with the largest logit, which is restricted to discrete grid points and can be unstable when neighboring voxels compete. A top-$k$ soft argmax decoder, instead, takes a probability-weighted average over the $k$ voxel centers with the largest logits, applied after a softmax with temperature $T$ over those specific $k$ logits:
\begin{equation}
    \hat{a}
    \;=\;
    \sum_{b \in \mathrm{TopK}_k(z)}
        \widetilde{\mathrm{softmax}}(z / T)_b \cdot b,
    \label{eq:topk_decode}
\end{equation}

\noindent where $\widetilde{\mathrm{softmax}}$ is the softmax renormalized over the $k$ retained voxels. The temperature $T$ controls the sharpness of this distribution. A smaller $T$ concentrates the average around the peak, while a larger $T$ spreads it across all $k$ voxels. We use $k{=}10$ and $T{=}1.0$ throughout the main experiments, and we ablate these choices in Section~\ref{sec:ablation_decoding}. Finally, the discrete gripper command is simply read from the argmax over its two bins.
\section{Experiments}

\subsection{Experimental Setup}

\subsubsection{Baselines}

We use two backbone VLAs as baselines: OpenVLA-OFT \cite{OpenVLA-OFT}, a Prismatic-7B VLM with a parallel-decoding L1-regression head, and $\pi_{0.5}$ \cite{Pi-0.5}, a VLM with a flow-matching action expert. Our heatmap action head replaces each backbone's action decoder, leaving the rest of the backbone recipe untouched, such as $\pi_{0.5}$'s flow-matching decoding mechanism. We use OpenVLA-OFT as the main backbone for both LIBERO and the real-world Franka studies. We use $\pi_{0.5}$ as a cross-backbone verification on LIBERO, with real-world Franka studies reported in Appendix~\ref{sec:appendix_pi_real_world}.

\subsubsection{LIBERO Experiments}
\label{sec:libero_experiments}

\paragraph{Tasks and evaluation.} We evaluate on the four LIBERO suites, Spatial, Object, Goal, Long \cite{LIBERO}, following OpenVLA-OFT's default input setup. Each success rate value reports 50 trials $\times$ 10 tasks $=$ 500 episodes at LIBERO seed 7. For the data efficiency studies, we additionally finetune models on stratified-per-task 10\%, 25\%, and 50\% subsets of LIBERO-Spatial. 

\paragraph{Finetuning.} All OpenVLA-OFT runs perform LoRA finetuning (rank 32, $\alpha=16$) on all linear layers for 10000 optimizer steps. We optimize using AdamW at a constant learning rate of $5 \times 10^{-4}$ with no warm-up and no weight decay. The effective batch size is 64, split across a 2 $\times$ H200 DDP setup with a per-GPU batch size of 8 and 4 gradient accumulation steps. Our heatmap head discretizes translation onto a $48 \times 48 \times 24$ grid and rotation onto a $24 \times 24 \times 24$ Euler grid. We train the heatmap head with cross-entropy against a Gaussian-blob target. This target uses $\sigma=0.1$ in normalized $[-1, 1]$ units for both translation and rotation grids. At inference, we decode by soft-argmax over the top 10 bins at a temperature of $T=1.0$. The OpenVLA-OFT baseline keeps every other hyperparameter fixed and uses its default L1 regression head. For $\pi_{0.5}$ experiments, we follow its public LIBERO recipe, with details provided in Appendix~\ref{sec:appendix_pi_libero}.

\subsubsection{Real-World Franka Experiments}
\label{sec:real_world_franka_setup}

\paragraph{Tasks and evaluation.} For OpenVLA-OFT experiments, we evaluate three real-world tasks on a Franka Research 3 robot:

\begin{enumerate}
    \item \textbf{Pick}: \textit{"Pick up the brick and place it on the green coaster"}; 225 demonstrations. 
    \item \textbf{Sweep}: \textit{"Push the gray brick into the white dustpan"}; 200 demonstrations. 
    \item \textbf{Insert}: \textit{"Insert the IC card into the orange card slot"}; 275 demonstrations. 
\end{enumerate}

We additionally use a 50-episode subset from each task to finetune models for data efficiency studies. During evaluation, we do 5 trials at object locations seen during training, and 5 at unseen locations.

\paragraph{Finetuning.} The setup follows Section \ref{sec:libero_experiments} except for the differences below. The voxel heatmap head uses a $48 \times 48 \times 48$ grid for translation and a $16 \times 16 \times 16$ grid for rotation. We calculate the 1\textsuperscript{st} and 99\textsuperscript{th} quantiles $q_{01}$ and $q_{99}$ from the training data for each action axis, and lay out the heatmap grid over $[q_{01} - 2\sigma, q_{99} + 2\sigma]$ per axis. We then normalize this padded range to the $[0, 1]$ grid coordinates, and use the same $\sigma=0.1$ as the Gaussian-blob target. We use front and wrist cameras as inputs. We train for 4000 steps, which is sufficient for empirical convergence at this dataset size. The OpenVLA-OFT baseline mirrors this setup.

\subsection{Quantitative Analysis}
\label{sec:exp_main}

\begin{figure}
    \centering
    \includegraphics[width=\linewidth]{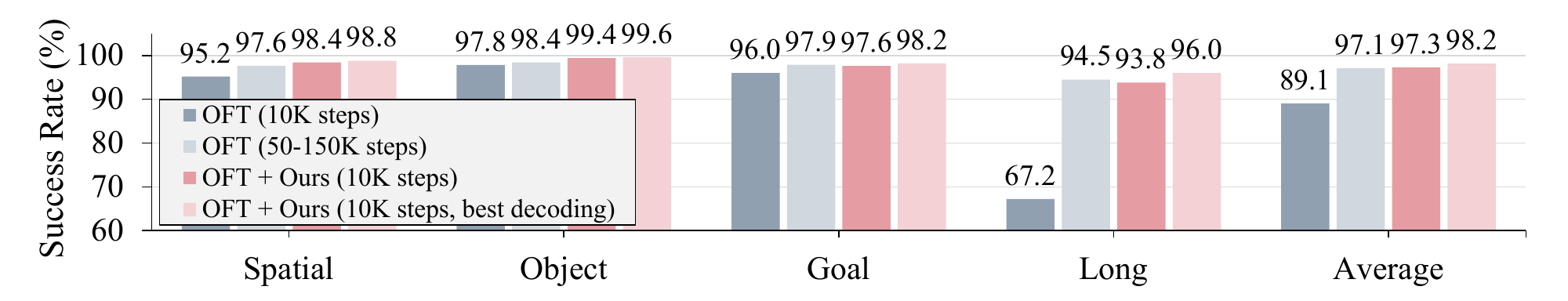}
    \includegraphics[width=\linewidth]{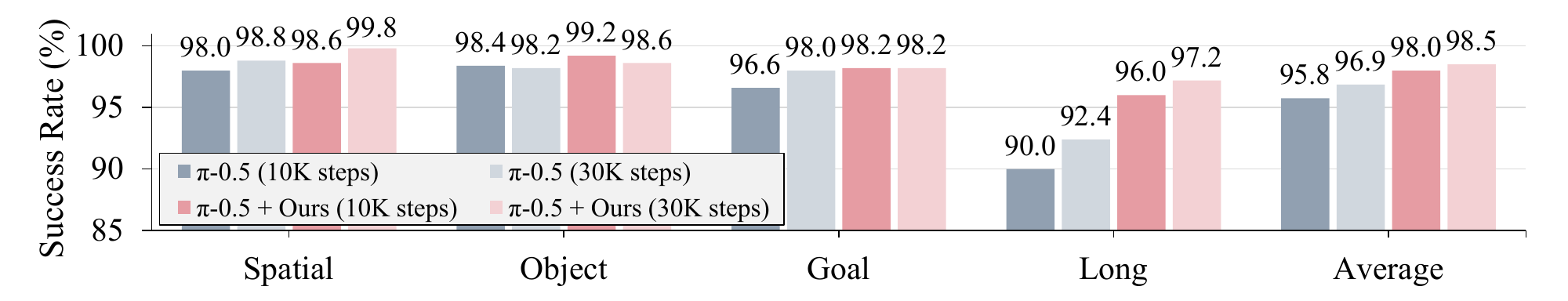}
    \caption{LIBERO main results across two backbones. \textbf{(top)} On OpenVLA-OFT, our heatmap head at 10K steps with the default decoder improves the four-suite average over the matched-step OpenVLA-OFT baseline, while our best decoders' results at 10K steps consistently outperform the published OpenVLA-OFT results (at 50K-150K training steps). \textbf{(bottom)} On $\pi_{0.5}$, our heatmap head at 30K steps outperforms the published $\pi_{0.5}$ baseline on all four suites; at 10K steps it already exceeds the 30K-step $\pi_{0.5}$ baseline on the four-suite average. On both backbones, the largest gain falls on LIBERO-Long, suggesting that our heatmap head is most beneficial on long-horizon tasks. OFT 50-150K and $\pi_{0.5}$ 30K bars report values from \cite{OpenVLA-OFT, Pi-0.5}.}
    \label{fig:libero_results_main}
\end{figure}

\paragraph{ActionMap improves LIBERO performance on two architecturally distinct backbones.} As shown in Fig.~\ref{fig:libero_results_main}, at matched training steps (10K steps for OpenVLA-OFT, 30K steps for $\pi_{0.5}$), our heatmap head improves the four-suite average by +8.2\% over OpenVLA-OFT's L1 regression head and by +1.6\% over $\pi_{0.5}$'s flow-matching head. On both backbones, the largest gain is in the hardest suite LIBERO-Long: +26.6\% on OpenVLA-OFT and +4.8\% on $\pi_{0.5}$, suggesting that our heatmap head is most beneficial on long-horizon tasks.

\paragraph{ActionMap matches or surpasses each backbone's published full-budget result at fewer or equal training steps.} Our 10000-step OpenVLA-OFT run with the default decoder of Section~\ref{sec:methodology_inference} already matches the published OpenVLA-OFT result (Fig.~\ref{fig:libero_results_main}, top) at 5 to 15 times less training, and surpasses it by +1.1\% under the per-suite best decoder among the alternatives in Section~\ref{sec:ablation_decoding}. This convergence advantage is most visible on LIBERO-Long, where our 10K-step run with best decoding already exceeds the published 50K- to 150K-step OpenVLA-OFT result. On $\pi_{0.5}$ (Fig.~\ref{fig:libero_results_main}, bottom), our heatmap head at 10K steps already surpasses the published 30K-step $\pi_{0.5}$ baseline on the four-suite average by +1.1\% at one-third the training steps; trained for the same 30K steps, the gain grows to +1.6\%.

\begin{figure}
    \centering
    \includegraphics[width=\linewidth]{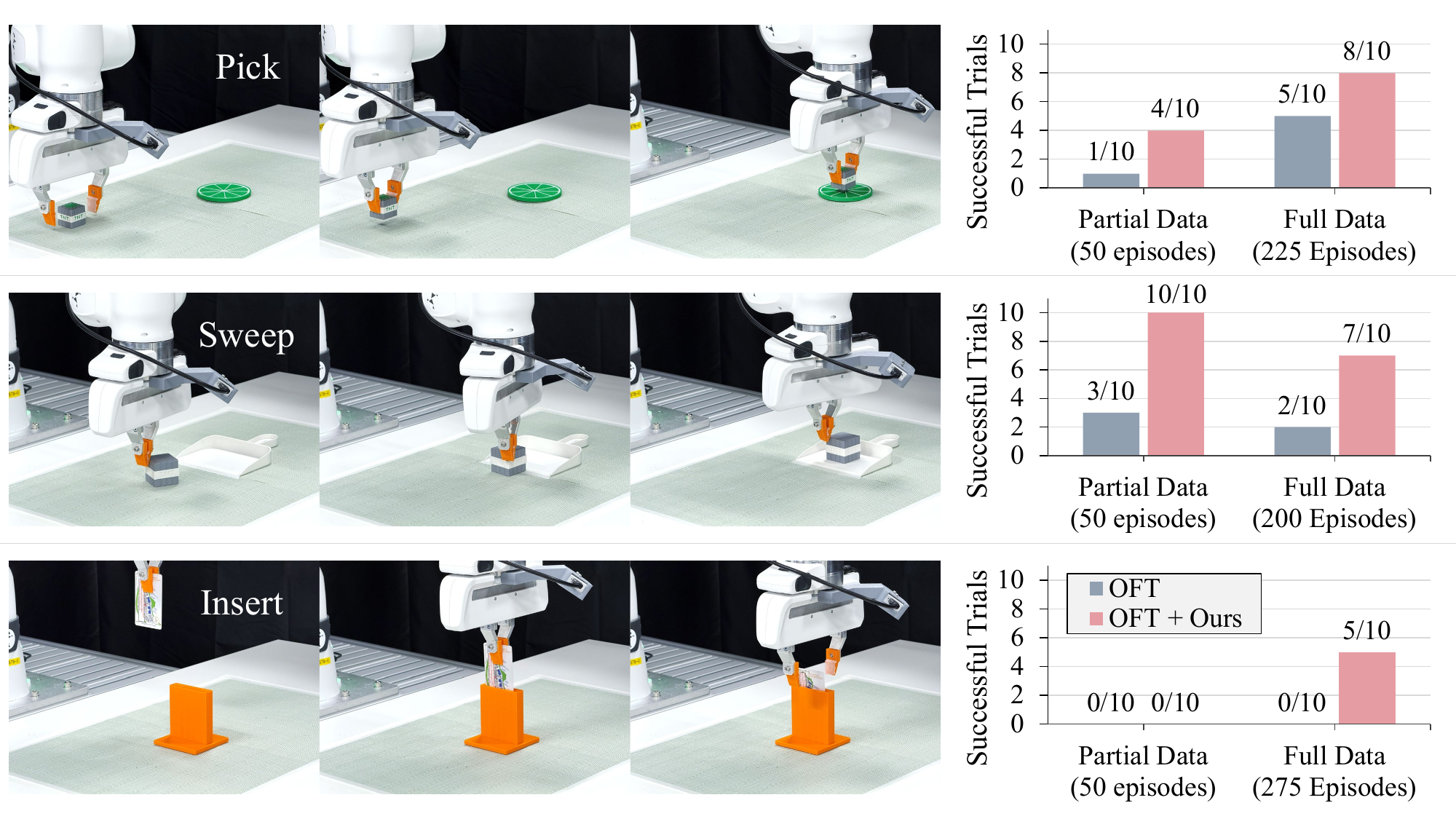}
    \caption{Real-world Franka results on three tasks: \textbf{Pick} (top), \textbf{Sweep} (middle), and \textbf{Insert} (bottom). For each task, the left three images show a representative successful rollout of our heatmap head, and the right bar chart reports successful trials at the partial-data setting and at full data. Our heatmap head outperforms OpenVLA-OFT's L1 regression head on every task at full data and on Pick and Sweep at the partial-data setting; on Insert at the partial-data setting, neither head succeeds.}
    \label{fig:real_world_results}
\end{figure}

\paragraph{ActionMap improves real-world Franka performance across all tasks and data scales.} On the three Franka tasks defined in Section~\ref{sec:real_world_franka_setup}, with 10 trials per cell, our heatmap head wins on every task at full data (pooled 20/30 vs.\ 7/30) over OpenVLA-OFT's L1 regression head, and on two of the three tasks at the 50-episode partial-data setting (pooled 14/30 vs.\ 4/30, Fig.~\ref{fig:real_world_results}). Both heads reach 0/10 on the Insert task at 50 episodes, suggesting that 50 demonstrations are insufficient for either head to learn Insert's two-stage behavior of first aligning the card above the slot, then inserting vertically down, which we analyze qualitatively in detail in Appendix~\ref{sec:appendix_qualitative}.

\begin{figure}
    \centering
    \includegraphics[width=0.41\linewidth]{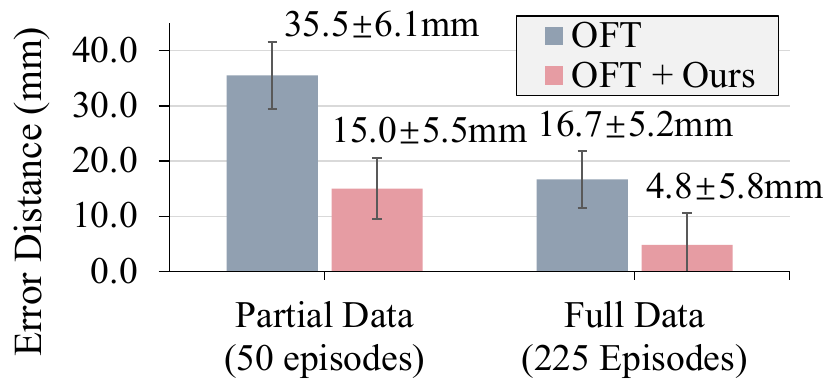}
    \caption{End-effector position error at the grasp moment on Pick (mean $\pm$ standard deviation, in mm). Our heatmap head produces tighter grasp localization than OpenVLA-OFT's L1 regression head at both data scales, and grasp error decreases for both heads as training data grows.}
    \label{fig:pick_pose_error}
\end{figure}

\paragraph{ActionMap's real-world advantage comes from more precise end-effector positioning.} On Pick, where the per-location dataset reference is documented, we measure the average end-effector position error at the grasp moment. As shown in Fig.~\ref{fig:pick_pose_error}, our heatmap head grasps within 4.8\,mm of the reference at full data and 15.0\,mm at the 50-episode setting, a 2 to 3 times tighter localization than OpenVLA-OFT's L1 regression head. This tighter localization at both data scales directly explains the higher Pick success rate, and is consistent with our heatmap head's distributional supervision learning the action space's spatial structure more accurately than point-estimate regression (see Appendix~\ref{sec:appendix_qualitative} for a per-trial failure-mode analysis). The corresponding $\pi_{0.5}$ real-world comparison is provided in Appendix~\ref{sec:appendix_pi_real_world}.

\begin{figure}
    \centering
    \includegraphics[width=\linewidth]{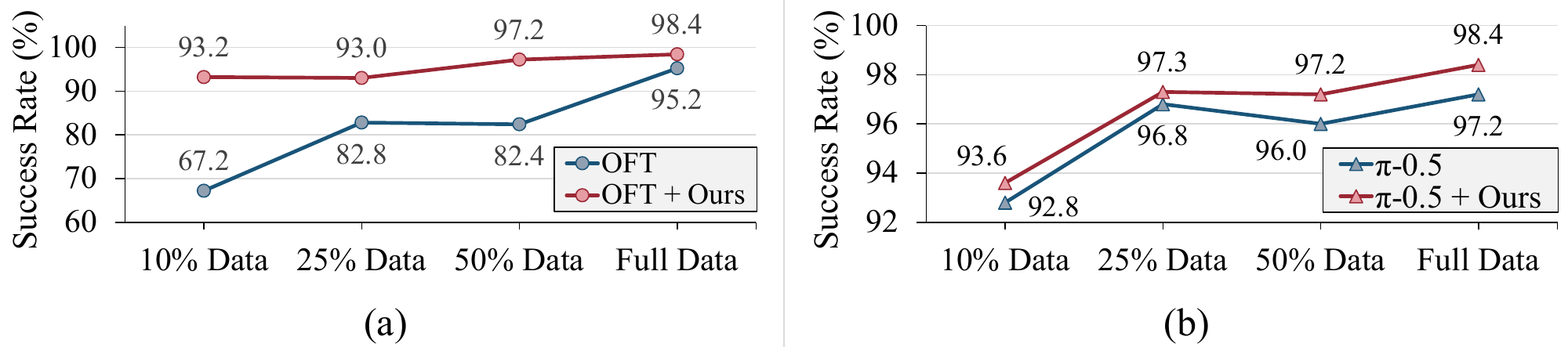}
    \caption{Data efficiency on LIBERO-Spatial across four training-data fractions for both backbones. \textbf{(a)} On OpenVLA-OFT, our heatmap head retains a substantial advantage at low data while the L1 regression head's success rate collapses. \textbf{(b)} On $\pi_{0.5}$, our heatmap head retains a small but consistent lead across all four fractions. The widening gap at low data on OpenVLA-OFT suggests that distributional supervision extracts more signal per demonstration than point-estimate regression.}
    \label{fig:libero_data_efficiency}
\end{figure}

\paragraph{ActionMap retains its advantage at lower training-data fractions.} We finetune both heads on stratified 10\%, 25\%, 50\%, and 100\% subsets of LIBERO-Spatial. As shown in Fig.~\ref{fig:libero_data_efficiency}(a), the gap on OpenVLA-OFT widens as the training set shrinks: at 10\% data (43 demonstrations) our heatmap head holds 93.2\% while OpenVLA-OFT's L1 regression head collapses to 67.2\%, a +26.0\% gap; at 50\% data our heatmap head already exceeds the L1 regression head's full-data score. On $\pi_{0.5}$ (Fig.~\ref{fig:libero_data_efficiency}(b)), both heads sit above 92\% across all four data fractions, with our heatmap head retaining a small consistent lead. The widening gap at low data shows that each demonstration is more efficiently exploited under distributional supervision over the entire grid than under point-estimate regression.


\begin{figure}
    \centering
    \includegraphics[width=\linewidth]{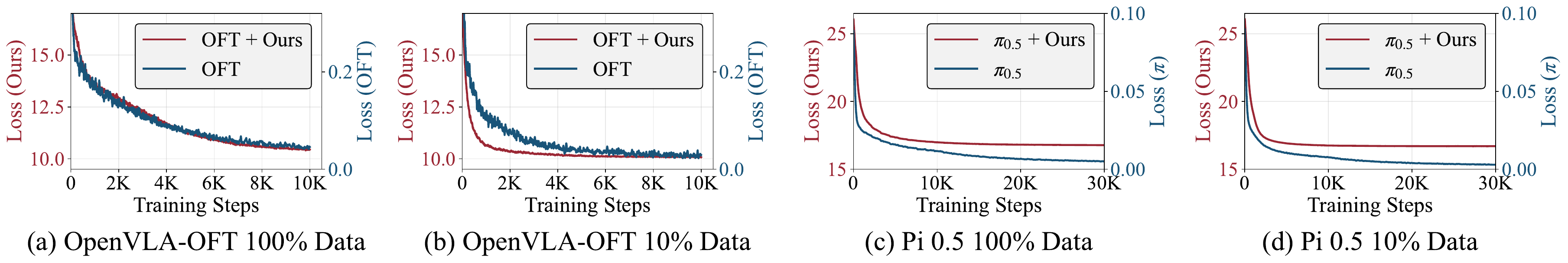}
    \caption{Training loss on LIBERO-Spatial across both backbones and two training-data fractions. \textbf{(a)} OpenVLA-OFT at 100\% data. \textbf{(b)} OpenVLA-OFT at 10\% data. \textbf{(c)} $\pi_{0.5}$ at 100\% data. \textbf{(d)} $\pi_{0.5}$ at 10\% data. Different loss functions are plotted against separate y-axes (red for our heatmap head, blue for the corresponding baseline). The red curve reaches its plateau no later than the blue curve in every panel, indicating that our heatmap head accelerates training convergence.}
    \label{fig:libero_loss}
\end{figure}

\paragraph{ActionMap accelerates training convergence.} Fig.~\ref{fig:libero_loss} plots training loss for both heads under each backbone at 100\% and 10\% LIBERO-Spatial data. The two heads minimize different loss functions, so their absolute loss values are not directly comparable; the meaningful comparison is the rate at which each curve plateaus. Across three (b,c,d) of all four panels, our heatmap head's loss reaches its plateau faster than the corresponding baseline's loss. On (a) convergence rates are comparable. The most striking gap is on OpenVLA-OFT at 10\% data, where the L1 regression head's loss continues to drop throughout the 10000-step run while our heatmap head's loss saturates within roughly 2k steps. This earlier saturation underlies our heatmap head's ability to reach the published full-budget result of each backbone at fewer or equal training steps. Another key observation is that our loss curve for OpenVLA-OFT is significantly smoother, suggesting a more stable learning process.

\subsection{Ablation Studies}
\label{sec:exp_ablation}

We perform two ablations on OpenVLA-OFT to characterize the design choices of our heatmap head: the inference-time decoding strategy, and the heatmap target hyperparameters of grid resolution and Gaussian blob standard deviation $\sigma$.

\subsubsection{Ablation on Action Decoding Strategy}
\label{sec:ablation_decoding}

\begin{figure}
    \centering
    \includegraphics[width=\linewidth]{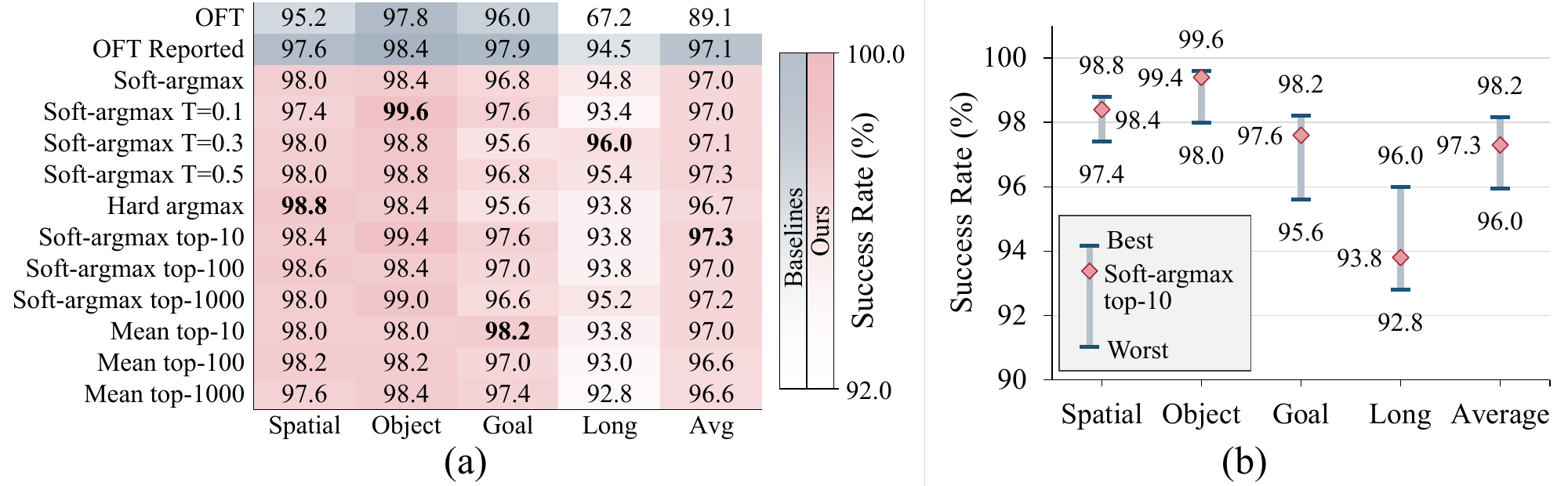}
    \caption{Ablation of inference-time decoding strategies on OpenVLA-OFT. \textbf{(a)} Per-suite success rates of two baselines (blue rows: OFT at 10K steps, OFT Reported at 50K-150K steps) and 11 alternative decoding strategies for our heatmap head (red rows). \textbf{(b)} Per-suite spread between the best and worst decoding strategy among the 11 alternatives (vertical blue bars), with our default decoder (\emph{Soft-argmax top-10}) marked as a red diamond. The 11 strategies span only 0.75\% on the four-suite average and the per-suite spread is at most 3.2\%, indicating that LIBERO performance under our heatmap head is largely insensitive to the decoder choice.}
    \label{fig:ablation_decoding}
\end{figure}

Our default decoder is a soft-argmax over the top-10 bins at temperature 1.0 (Section~\ref{sec:methodology_inference}). To test sensitivity to this choice, we evaluate ten alternative decoders on the same OpenVLA-OFT checkpoints: soft-argmax over the full grid at temperatures 0.1, 0.3, 0.5, and 1.0; soft-argmax over the top-100 and top-1000 bins; the unweighted mean of the top-10, top-100, and top-1000 bin centers (a uniform average over the selected bins, with no softmax weighting); and hard argmax. As shown in Fig.~\ref{fig:ablation_decoding}, decoder choice has little impact overall: the eleven four-suite averages span less than 1\%, with LIBERO-Long the most decoder-sensitive suite at a spread of 3.2\%.

\subsubsection{Ablation on Grid Resolution and Gaussian Blob Width}
\label{sec:ablation_grid_sigma}

\begin{figure}
    \centering
    \includegraphics[width=0.67\linewidth]{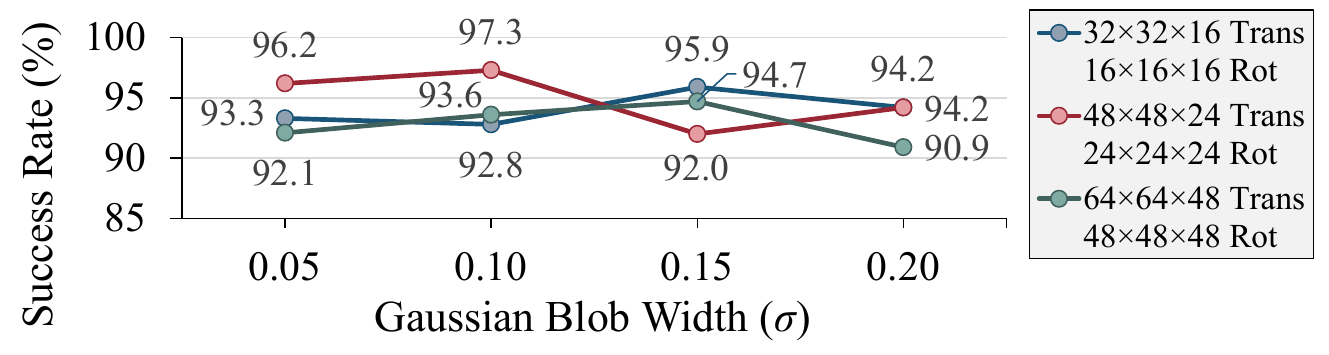}
    \caption{Ablation of grid resolution and Gaussian-blob width on LIBERO. Each value in the plot is an average across all four LIBERO subsets. Three grid resolutions (translation paired with rotation, increasing in density) crossed with four Gaussian-blob widths. Performance across the twelve cells stays within a 6.4\% band, with our headline configuration (red, $\sigma=0.10$) reaching the global optimum, indicating broad robustness to both axes around the headline.}
    \label{fig:ablation_grid_sigma}
\end{figure}

The heatmap target distribution is parameterized by two hyperparameters: the grid resolution and the Gaussian blob standard deviation $\sigma$. We sweep three grid resolutions of increasing density (translation $32{\times}32{\times}16$ paired with rotation $16{\times}16{\times}16$; translation $48{\times}48{\times}24$ paired with rotation $24{\times}24{\times}24$; translation $64{\times}64{\times}48$ paired with rotation $48{\times}48{\times}48$) crossed with four $\sigma$ values (0.05, 0.10, 0.15, 0.20) across the four LIBERO suites, to characterize how the target-distribution width interacts with grid resolution. As shown in Fig.~\ref{fig:ablation_grid_sigma}, performance across the twelve cells stays within a 6.4\% band, with our headline configuration at $\sigma=0.10$ reaching the global optimum. We draw two observations: (i) for each grid, the optimal $\sigma$ lies in a moderate range; and (ii) the same holds for grid resolution: both coarser and finer grids underperform our headline.

\section{Conclusion}


We presented \textbf{ActionMap}, a voxel heatmap action head that drops into an existing VLA, trained against a Gaussian-blob target via cross-entropy. Validated on OpenVLA-OFT (L1 regression) and $\pi_{0.5}$ (flow matching) in LIBERO and on real-world Franka manipulation, it yields gains in success rate, training convergence, and data efficiency, suggesting that action representation is itself a meaningful design axis for VLA performance.

\textbf{Limitations and Future Analysis.} The voxel grid (and the corresponding network parameters in the heatmap head) scales polynomially with per-axis resolution, significantly limiting the grid granularity we can practically employ; addressing this naturally motivates \emph{adaptive grid resolutions} that allocate density where it is most needed. Beyond this, our heatmap action head opens several other future directions: \emph{absolute-coordinate grids} as an alternative to the delta-action layout, \emph{adaptive supervision} that varies the Gaussian-blob width across grid resolutions, tasks, or training stages, and \emph{multimodal or temporal sampling} that harnesses the full distributional output beyond per-frame argmax. We elaborate on each direction in Appendix~\ref{sec:appendix_future}.




{
\small
\bibliographystyle{IEEEtran}
\bibliography{references}

@inproceedings{OpenVLA,
  title     = {OpenVLA: An Open-Source Vision-Language-Action Model},
  author    = {Kim, Moo Jin and Pertsch, Karl and Karamcheti, Siddharth and Xiao, Ted and Balakrishna, Ashwin and Nair, Suraj and Rafailov, Rafael and Foster, Ethan and Lam, Grace and Sanketi, Pannag and Vuong, Quan and Kollar, Thomas and Burchfiel, Benjamin and Tedrake, Russ and Sadigh, Dorsa and Levine, Sergey and Liang, Percy and Finn, Chelsea},
  booktitle = {Conference on Robot Learning (CoRL)},
  year      = {2024},
  eprint    = {2406.09246},
  archivePrefix = {arXiv}
}

@inproceedings{OpenVLA-OFT,
  title     = {Fine-Tuning Vision-Language-Action Models: Optimizing Speed and Success},
  author    = {Kim, Moo Jin and Finn, Chelsea and Liang, Percy},
  booktitle = {Robotics: Science and Systems (RSS)},
  year      = {2025},
  eprint    = {2502.19645},
  archivePrefix = {arXiv}
}

@article{Pi-0.5,
  title   = {{$\pi_{0.5}$}: A Vision-Language-Action Model with Open-World Generalization},
  author  = {Black, Kevin and Brown, Noah and Driess, Danny and Esmail, Adnan and Equi, Michael and Finn, Chelsea and Fusai, Niccolo and Galliker, Manuel Y. and Ghosh, Dibya and Groom, Lachy and Hausman, Karol and Ichter, Brian and Jakubczak, Szymon and Jones, Tim and Ke, Liyiming and LeBlanc, Devin and Levine, Sergey and Lin, Adrian and Mees, Oier and Pertsch, Karl and Sanketi, Pannag and Schaal, Stefan and Shi, Lucy Xiaoyang and Smith, Laura and Springenberg, Jost Tobias and Stone, Kyle and Tanner, James and Vuong, Quan and Walling, Anna and Wang, Haohuan and Welander, Joseph and Zhilinsky, Ury},
  journal = {arXiv preprint arXiv:2504.16054},
  year    = {2025}
}

@inproceedings{LIBERO,
  title     = {{LIBERO}: Benchmarking Knowledge Transfer for Lifelong Robot Learning},
  author    = {Liu, Bo and Zhu, Yifeng and Gao, Chongkai and Feng, Yihao and Liu, Qiang and Zhu, Yuke and Stone, Peter},
  booktitle = {Advances in Neural Information Processing Systems (NeurIPS)},
  year      = {2023},
  eprint    = {2306.03310},
  archivePrefix = {arXiv}
}

@article{zheng2025x,
  title={X-vla: Soft-prompted transformer as scalable cross-embodiment vision-language-action model},
  author={Zheng, Jinliang and Li, Jianxiong and Wang, Zhihao and Liu, Dongxiu and Kang, Xirui and Feng, Yuchun and Zheng, Yinan and Zou, Jiayin and Chen, Yilun and Zeng, Jia and others},
  journal={arXiv preprint arXiv:2510.10274},
  year={2025}
}

@inproceedings{brohan2023rt2,
  title     = {{RT-2}: Vision-Language-Action Models Transfer Web Knowledge
               to Robotic Control},
  author    = {Brohan, Anthony and Brown, Noah and Carbajal, Justice and
               Chebotar, Yevgen and Chen, Xi and Choromanski, Krzysztof and
               Ding, Tianli and Driess, Danny and Dubey, Kumar Avinava and
               Finn, Chelsea and Florence, Pete and Fu, Chuyuan and
               Gopalakrishnan, Keerthana and Hausman, Karol and Herzog,
               Alexander and Hsu, Jasmine and Ichter, Brian and Irpan, Alex
               and Joshi, Nikhil J. and Julian, Ryan and Kalashnikov, Dmitry
               and Kuang, Yuheng and Leal, Isabel and Levine, Sergey and
               Lu, Yao and Michalewski, Henryk and Mordatch, Igor and
               Pertsch, Karl and Rao, Kanishka and Reymann, Krista and
               Ryoo, Michael S. and Salazar, Grecia and Sanketi, Pannag and
               Sermanet, Pierre and Singh, Anikait and Singh, Jaspiar and
               Soricut, Radu and Tran, Huong and Vanhoucke, Vincent and
               Vuong, Quan and Wahid, Ayzaan and Welker, Stefan and
               Wohlhart, Paul and Wu, Jialin and Xiao, Ted and Xu, Peng and
               Xu, Sichun and Yu, Tianhe and Zitkovich, Brianna},
  booktitle = {Proceedings of The 7th Conference on Robot Learning},
  pages     = {2165--2183},
  volume    = {229},
  series    = {Proceedings of Machine Learning Research},
  year      = {2023}
}

@article{black2024pi0,
  title   = {{$\pi_0$}: A Vision-Language-Action Flow Model for General
             Robot Control},
  author  = {Black, Kevin and Brown, Noah and Driess, Danny and Esmail,
             Adnan and Equi, Michael and Finn, Chelsea and Fusai, Niccolo
             and Groom, Lachy and Hausman, Karol and Ichter, Brian and
             Jakubczak, Szymon and Jones, Tim and Ke, Liyiming and Levine,
             Sergey and Li-Bell, Adrian and Mothukuri, Mohith and Nair,
             Suraj and Pertsch, Karl and Shi, Lucy Xiaoyang and Tanner,
             James and Vuong, Quan and Walling, Anna and Wang, Haohuan and
             Zheng, Ury},
  journal = {arXiv preprint arXiv:2410.24164},
  year    = {2024}
}

@article{gr00tn1_2025,
  title   = {{GR00T N1}: An Open Foundation Model for Generalist
             Humanoid Robots},
  author  = {Bjorck, Johan and Casta{\~n}eda, Fernando and Cherniadev,
             Nikita and Da, Xingye and Ding, Runyu and Fan, Linxi and
             Fang, Yu and Fox, Dieter and Hu, Fengyuan and Huang, Spencer
             and Jang, Joel and Jiang, Zhenyu and Kautz, Jan and Kundalia,
             Kaushil and Lao, Lawrence and Li, Zhiqi and Lin, Zongyu and
             Lin, Kevin and Liu, Guilin and Llontop, Edith and Magne, Loic
             and Mandlekar, Ajay and Narayan, Avnish and Nasiriany, Soroush
             and Reed, Scott and Tan, You Liang and Wang, Guanzhi and
             Wang, Zu and Wang, Jing and Wang, Qi and Xiang, Jiannan and
             Xie, Yuqi and Xu, Yinzhen and Xu, Zhenjia and Ye, Seonghyeon
             and Yu, Zhiding and Zhang, Ao and Zhang, Hao and Zhao, Yizhou
             and Zheng, Ruijie and Zhu, Yuke},
  journal = {arXiv preprint arXiv:2503.14734},
  year    = {2025}
}

@inproceedings{chi2023diffusion,
  title     = {Diffusion Policy: Visuomotor Policy Learning via Action
               Diffusion},
  author    = {Chi, Cheng and Feng, Siyuan and Du, Yilun and Xu, Zhenjia
               and Cousineau, Eric and Burchfiel, Benjamin and Song, Shuran},
  booktitle = {Proceedings of Robotics: Science and Systems (RSS)},
  year      = {2023},
  doi       = {10.15607/RSS.2023.XIX.016}
}

@inproceedings{zhao2023act,
  title     = {Learning Fine-Grained Bimanual Manipulation with Low-Cost
               Hardware},
  author    = {Zhao, Tony Z. and Kumar, Vikash and Levine, Sergey and
               Finn, Chelsea},
  booktitle = {Proceedings of Robotics: Science and Systems (RSS)},
  year      = {2023},
  doi       = {10.15607/RSS.2023.XIX.016}
}

@inproceedings{shridhar2022perceiver,
  title     = {Perceiver-Actor: A Multi-Task Transformer for Robotic
               Manipulation},
  author    = {Shridhar, Mohit and Manuelli, Lucas and Fox, Dieter},
  booktitle = {Proceedings of the 6th Conference on Robot Learning (CoRL)},
  pages     = {785--799},
  volume    = {205},
  series    = {Proceedings of Machine Learning Research},
  year      = {2022}
}

@inproceedings{goyal2023rvt,
  title     = {{RVT}: Robotic View Transformer for 3D Object Manipulation},
  author    = {Goyal, Ankit and Xu, Jie and Guo, Yijie and Blukis, Valts
               and Chao, Yu-Wei and Fox, Dieter},
  booktitle = {Proceedings of the 7th Conference on Robot Learning (CoRL)},
  series    = {Proceedings of Machine Learning Research},
  volume    = {229},
  year      = {2023}
}

@inproceedings{goyal2024rvt2,
  title     = {{RVT-2}: Learning Precise Manipulation from Few
               Demonstrations},
  author    = {Goyal, Ankit and Blukis, Valts and Xu, Jie and Guo, Yijie
               and Chao, Yu-Wei and Fox, Dieter},
  booktitle = {Proceedings of Robotics: Science and Systems (RSS)},
  year      = {2024}
}

@inproceedings{huang2023voxposer,
  title     = {{VoxPoser}: Composable 3D Value Maps for Robotic
               Manipulation with Language Models},
  author    = {Huang, Wenlong and Wang, Chen and Zhang, Ruohan and Li,
               Yunzhu and Wu, Jiajun and Fei-Fei, Li},
  booktitle = {Proceedings of the 7th Conference on Robot Learning (CoRL)},
  series    = {Proceedings of Machine Learning Research},
  volume    = {229},
  year      = {2023}
}

@article{li2025bridgevla,
  title   = {{BridgeVLA}: Input-Output Alignment for Efficient 3D
             Manipulation Learning with Vision-Language Models},
  author  = {Li, Peiyan and Chen, Yixiang and Wu, Hongtao and Ma, Xiao
             and Wu, Xiangnan and Huang, Yan and Wang, Liang and Kong, Tao
             and Tan, Tieniu},
  journal = {arXiv preprint arXiv:2506.07961},
  year    = {2025}
}

@inproceedings{sun2018integral,
  title     = {Integral Human Pose Regression},
  author    = {Sun, Xiao and Xiao, Bin and Wei, Fangyin and Liang, Shuang
               and Wei, Yichen},
  booktitle = {Proceedings of the European Conference on Computer Vision
               (ECCV)},
  pages     = {529--545},
  year      = {2018}
}

@inproceedings{rt1,
  title     = {{RT-1}: Robotics Transformer for Real-World Control at Scale},
  author    = {Brohan, Anthony and Brown, Noah and Carbajal, Justice and
               Chebotar, Yevgen and Dabis, Joseph and Finn, Chelsea and
               Gopalakrishnan, Keerthana and Hausman, Karol and Herzog,
               Alex and Hsu, Jasmine and others},
  booktitle = {Robotics: Science and Systems (RSS)},
  year      = {2023},
  eprint    = {2212.06817},
  archivePrefix = {arXiv}
}

@inproceedings{octo,
  title     = {{Octo}: An Open-Source Generalist Robot Policy},
  author    = {{Octo Model Team} and Ghosh, Dibya and Walke, Homer and
               Pertsch, Karl and Black, Kevin and Mees, Oier and Dasari,
               Sudeep and Hejna, Joey and Xu, Charles and Luo, Jianlan and
               others},
  booktitle = {Robotics: Science and Systems (RSS)},
  year      = {2024},
  eprint    = {2405.12213},
  archivePrefix = {arXiv}
}

@inproceedings{roboflamingo,
  title     = {Vision-Language Foundation Models as Effective Robot Imitators},
  author    = {Li, Xinghang and Liu, Minghuan and Zhang, Hanbo and Yu,
               Cunjun and Xu, Jie and Wu, Hongtao and Cheang, Chilam and
               Jing, Ya and Zhang, Weinan and Liu, Huaping and Li, Hang and
               Kong, Tao},
  booktitle = {International Conference on Learning Representations (ICLR)},
  year      = {2024},
  eprint    = {2311.01378},
  archivePrefix = {arXiv}
}

@inproceedings{hpt,
  title     = {Scaling Proprioceptive-Visual Learning with Heterogeneous
               Pre-trained Transformers},
  author    = {Wang, Lirui and Chen, Xinlei and Zhao, Jialiang and He,
               Kaiming},
  booktitle = {Advances in Neural Information Processing Systems (NeurIPS)},
  year      = {2024},
  eprint    = {2409.20537},
  archivePrefix = {arXiv}
}

@inproceedings{rdt,
  title     = {{RDT-1B}: A Diffusion Foundation Model for Bimanual Manipulation},
  author    = {Liu, Songming and Wu, Lingxuan and Li, Bangguo and Tan,
               Hengkai and Chen, Huayu and Wang, Zhengyi and Xu, Ke and
               Su, Hang and Zhu, Jun},
  booktitle = {International Conference on Learning Representations (ICLR)},
  year      = {2025},
  eprint    = {2410.07864},
  archivePrefix = {arXiv}
}

@article{tinyvla,
  title   = {{TinyVLA}: Toward Fast, Data-Efficient Vision-Language-Action
             Models for Robotic Manipulation},
  author  = {Wen, Junjie and Zhu, Yichen and Li, Jinming and Zhu, Minjie and
             Wu, Kun and Xu, Zhiyuan and Liu, Ning and Cheng, Ran and
             Shen, Chaomin and Peng, Yaxin and Feng, Feifei and Tang,
             Jian},
  journal = {IEEE Robotics and Automation Letters (RA-L)},
  year    = {2025},
  eprint  = {2409.12514},
  archivePrefix = {arXiv}
}

@inproceedings{droid,
  title     = {{DROID}: A Large-Scale In-the-Wild Robot Manipulation Dataset},
  author    = {Khazatsky, Alexander and Pertsch, Karl and Nair, Suraj and
               Balakrishna, Ashwin and Dasari, Sudeep and Karamcheti,
               Siddharth and Nasiriany, Soroush and Srirama, Mohan Kumar and
               Chen, Lawrence Yunliang and Ellis, Kirsty and others},
  booktitle = {Robotics: Science and Systems (RSS)},
  year      = {2024},
  eprint    = {2403.12945},
  archivePrefix = {arXiv}
}

@inproceedings{bridgev2,
  title     = {{BridgeData V2}: A Dataset for Robot Learning at Scale},
  author    = {Walke, Homer Rich and Black, Kevin and Zhao, Tony Z. and
               Vuong, Quan and Zheng, Chongyi and Hansen-Estruch, Philippe
               and He, Andre Wang and Myers, Vivek and Kim, Moo Jin and
               Du, Max and Lu, Abraham and Finn, Chelsea and Levine, Sergey},
  booktitle = {Conference on Robot Learning (CoRL)},
  year      = {2023},
  eprint    = {2308.12952},
  archivePrefix = {arXiv}
}

@inproceedings{robocasa,
  title     = {{RoboCasa}: Large-Scale Simulation of Everyday Tasks for
               Generalist Robots},
  author    = {Nasiriany, Soroush and Maddukuri, Abhiram and Zhang, Lance
               and Parikh, Adeet and Lo, Aaron and Joshi, Abhishek and
               Mandlekar, Ajay and Zhu, Yuke},
  booktitle = {Robotics: Science and Systems (RSS)},
  year      = {2024},
  eprint    = {2406.02523},
  archivePrefix = {arXiv}
}

@article{calvin,
  title   = {{CALVIN}: A Benchmark for Language-Conditioned Policy Learning
             for Long-Horizon Robot Manipulation Tasks},
  author  = {Mees, Oier and Hermann, Lukas and Rosete-Beas, Erick and
             Burgard, Wolfram},
  journal = {IEEE Robotics and Automation Letters (RA-L)},
  volume  = {7},
  number  = {3},
  pages   = {7327--7334},
  year    = {2022},
  eprint  = {2112.03227},
  archivePrefix = {arXiv}
}

@inproceedings{toshev2014deeppose,
  title     = {{DeepPose}: Human Pose Estimation via Deep Neural Networks},
  author    = {Toshev, Alexander and Szegedy, Christian},
  booktitle = {IEEE Conference on Computer Vision and Pattern Recognition (CVPR)},
  year      = {2014},
  eprint    = {1312.4659},
  archivePrefix = {arXiv}
}

@inproceedings{wei2016convolutional,
  title     = {Convolutional Pose Machines},
  author    = {Wei, Shih-En and Ramakrishna, Varun and Kanade, Takeo and
               Sheikh, Yaser},
  booktitle = {IEEE Conference on Computer Vision and Pattern Recognition (CVPR)},
  year      = {2016},
  eprint    = {1602.00134},
  archivePrefix = {arXiv}
}

@inproceedings{newell2016stacked,
  title     = {Stacked Hourglass Networks for Human Pose Estimation},
  author    = {Newell, Alejandro and Yang, Kaiyu and Deng, Jia},
  booktitle = {European Conference on Computer Vision (ECCV)},
  year      = {2016},
  eprint    = {1603.06937},
  archivePrefix = {arXiv}
}

@inproceedings{sun2019hrnet,
  title     = {Deep High-Resolution Representation Learning for Human Pose Estimation},
  author    = {Sun, Ke and Xiao, Bin and Liu, Dong and Wang, Jingdong},
  booktitle = {IEEE/CVF Conference on Computer Vision and Pattern Recognition (CVPR)},
  year      = {2019},
  eprint    = {1902.09212},
  archivePrefix = {arXiv}
}

@inproceedings{xiao2018simple,
  title={Simple baselines for human pose estimation and tracking},
  author={Xiao, Bin and Wu, Haiping and Wei, Yichen},
  booktitle={Proceedings of the European conference on computer vision (ECCV)},
  pages={466--481},
  year={2018}
}

@inproceedings{james2022coarse,
  title     = {Coarse-to-Fine {Q}-attention: Efficient Learning for Visual
               Robotic Manipulation via Discretisation},
  author    = {James, Stephen and Wada, Kentaro and Laidlow, Tristan and
               Davison, Andrew J.},
  booktitle = {IEEE/CVF Conference on Computer Vision and Pattern Recognition (CVPR)},
  year      = {2022},
  eprint    = {2106.12534},
  archivePrefix = {arXiv}
}

@inproceedings{ma2021context,
  title={Context modeling in 3d human pose estimation: A unified perspective},
  author={Ma, Xiaoxuan and Su, Jiajun and Wang, Chunyu and Ci, Hai and Wang, Yizhou},
  booktitle={Proceedings of the IEEE/CVF conference on computer vision and pattern recognition},
  pages={6238--6247},
  year={2021}
}

@article{padalkar2023oxe,
  title   = {Open {X-Embodiment}: Robotic Learning Datasets and
             {RT-X} Models},
  author  = {Padalkar, Abhishek and Pooley, Acorn and Jain, Ajinkya and
             others},
  journal = {arXiv preprint arXiv:2310.08864},
  year    = {2023}
}

@article{wu2023gr1,
  title   = {Unleashing Large-Scale Video Generative Pre-training for Visual Robot Manipulation},
  author  = {Wu, Hongtao and Jing, Ya and Cheang, Chilam and Chen, Guangzeng and Xu, Jiafeng and Li, Xinghang and Liu, Minghuan and Li, Hang and Kong, Tao},
  journal = {arXiv preprint arXiv:2312.13139},
  year    = {2023}
}

@article{cheang2024gr2,
  title   = {{GR-2}: A Generative Video-Language-Action Model with Web-Scale Knowledge for Robot Manipulation},
  author  = {Cheang, Chi-Lam and Chen, Guangzeng and Jing, Ya and Kong, Tao and Li, Hang and Li, Yifeng and Liu, Yuxiao and Wu, Hongtao and Xu, Jiafeng and Yang, Yichu and Zhang, Hanbo and Zhu, Minzhao},
  journal = {arXiv preprint arXiv:2410.06158},
  year    = {2024}
}

@article{cosmos_policy,
  title   = {Cosmos Policy: Fine-Tuning Video Models for Visuomotor Control and Planning},
  author  = {Kim, Moo Jin and Gao, Yihuai and Lin, Tsung-Yi and Lin, Yen-Chen and Ge, Yunhao and Lam, Grace and Liang, Percy and Song, Shuran and Liu, Ming-Yu and Finn, Chelsea and Gu, Jinwei},
  journal = {arXiv preprint arXiv:2601.16163},
  year    = {2026}
}

@article{dreamzero,
  title   = {World Action Models are Zero-shot Policies},
  author  = {Ye, Seonghyeon and Ge, Yunhao and Zheng, Kaiyuan and Gao, Shenyuan and Yu, Sihyun and Kurian, George and Indupuru, Suneel and Tan, You Liang and Zhu, Chuning and Xiang, Jiannan and Malik, Ayaan and Lee, Kyungmin and Liang, William and Ranawaka, Nadun and Gu, Jiasheng and Xu, Yinzhen and Wang, Guanzhi and Hu, Fengyuan and Narayan, Avnish and Bjorck, Johan and Wang, Jing and Kim, Gwanghyun and Niu, Dantong and Zheng, Ruijie and Xie, Yuqi and Wu, Jimmy and Wang, Qi and Julian, Ryan and Xu, Danfei and Du, Yilun and Chebotar, Yevgen and Reed, Scott and Kautz, Jan and Zhu, Yuke and Fan, Linxi Jim and Jang, Joel},
  journal = {arXiv preprint arXiv:2602.15922},
  year    = {2026}
}

@article{fastwam,
  title   = {{Fast-WAM}: Do World Action Models Need Test-time Future Imagination?},
  author  = {Yuan, Tianyuan and Dong, Zibin and Liu, Yicheng and Zhao, Hang},
  journal = {arXiv preprint arXiv:2603.16666},
  year    = {2026}
}

@inproceedings{ringid,
  title     = {{RingID}: Rethinking Tree-Ring Watermarking for Enhanced Multi-Key Identification},
  author    = {Ci, Hai and Yang, Pei and Song, Yiren and Shou, Mike Zheng},
  booktitle = {European Conference on Computer Vision (ECCV)},
  year      = {2024},
  eprint    = {2404.14055},
  archivePrefix = {arXiv}
}

@inproceedings{wmadapter,
  title     = {{WMAdapter}: Adding {WaterMark} Control to Latent Diffusion Models},
  author    = {Ci, Hai and Song, Yiren and Yang, Pei and Xie, Jinheng and Shou, Mike Zheng},
  booktitle = {International Conference on Machine Learning (ICML)},
  year      = {2025},
  eprint    = {2406.08337},
  archivePrefix = {arXiv}
}

@inproceedings{simple_avg_watermarks,
  title     = {Can Simple Averaging Defeat Modern Watermarks?},
  author    = {Yang, Pei and Ci, Hai and Song, Yiren and Shou, Mike Zheng},
  booktitle = {Advances in Neural Information Processing Systems (NeurIPS)},
  year      = {2024}
}

@inproceedings{idprotector,
  title     = {{IDProtector}: An Adversarial Noise Encoder to Protect Against ID-Preserving Image Generation},
  author    = {Song, Yiren and Yang, Pei and Ci, Hai and Shou, Mike Zheng},
  booktitle = {IEEE/CVF Conference on Computer Vision and Pattern Recognition (CVPR)},
  year      = {2025},
  eprint    = {2412.11638},
  archivePrefix = {arXiv}
}

@inproceedings{macosworld,
  title     = {{macOSWorld}: A Multilingual Interactive Benchmark for {GUI} Agents},
  author    = {Yang, Pei and Ci, Hai and Shou, Mike Zheng},
  booktitle = {Advances in Neural Information Processing Systems (NeurIPS)},
  year      = {2025},
  eprint    = {2506.04135},
  archivePrefix = {arXiv}
}

@misc{incontext_defense,
      title={In-Context Defense in Computer Agents: An Empirical Study}, 
      author={Pei Yang and Hai Ci and Mike Zheng Shou},
      year={2025},
      eprint={2503.09241},
      archivePrefix={arXiv},
      primaryClass={cs.AI},
      url={https://arxiv.org/abs/2503.09241}, 
}

@article{xhumanoid,
  title   = {{X-Humanoid}: Robotize Human Videos to Generate Humanoid Videos at Scale},
  author  = {Yang, Pei and Ci, Hai and Song, Yiren and Shou, Mike Zheng},
  journal = {arXiv preprint arXiv:2512.04537},
  year    = {2025}
}

@article{h2rgrounder,
  title   = {{H2R-Grounder}: A Paired-Data-Free Paradigm for Translating Human Interaction Videos into Physically Grounded Robot Videos},
  author  = {Ci, Hai and Liu, Xiaokang and Yang, Pei and Song, Yiren and Shou, Mike Zheng},
  journal = {arXiv preprint arXiv:2512.09406},
  year    = {2025}
}

@article{uenr600k,
  title   = {{UENR-600K}: A Large-Scale Physically Grounded Dataset for Nighttime Video Deraining},
  author  = {Yang, Pei and Ci, Hai and Lin, Beibei and Song, Yiren and Shou, Mike Zheng},
  journal = {arXiv preprint arXiv:2604.04402},
  year    = {2026}
}
}


\appendix
\section{OpenVLA-OFT Integration Details}
\label{sec:appendix_oft_integration}

\begin{figure}[h]
    \centering
    \includegraphics[width=\linewidth]{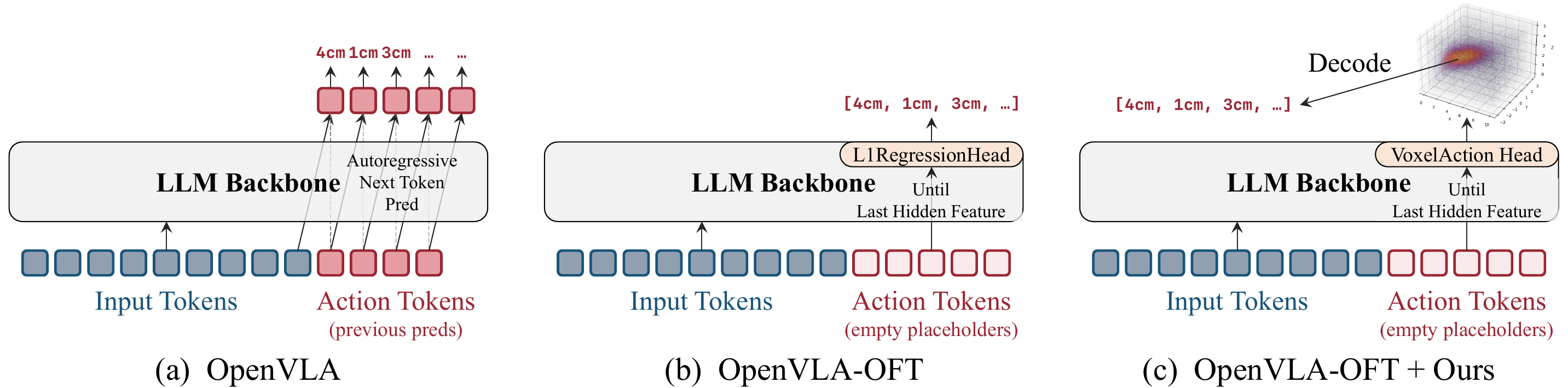}
    \caption{Integration of our voxel heatmap action head into OpenVLA-OFT, alongside the two prior decoder designs. \textbf{(a)} OpenVLA \cite{OpenVLA} predicts each action dimension autoregressively as a discrete language token. \textbf{(b)} OpenVLA-OFT \cite{OpenVLA-OFT} replaces the autoregressive token prediction with an L1 regression head that maps the last hidden feature of each action token directly to the continuous action vector. \textbf{(c)} Our voxel heatmap head reads the same last hidden features and outputs three voxel heatmaps (translation, rotation, gripper), from which the next action vector is decoded.}
    \label{fig:oft_integration}
\end{figure}

We integrate our voxel heatmap action head into OpenVLA-OFT \cite{OpenVLA-OFT} by replacing its L1 regression head while preserving the rest of the backbone, including the bidirectional attention over action-token positions and the parallel-decoding output of $T$ actions per forward pass. Fig.~\ref{fig:oft_integration} contrasts the resulting integration against the prior autoregressive OpenVLA \cite{OpenVLA} and the OpenVLA-OFT decoder.

\section{Implementation Details of $\pi_{0.5}$ LIBERO Experiments}
\label{sec:appendix_pi_libero}

\paragraph{Tasks and Evaluation.} We evaluate $\pi_{0.5}$ on the four LIBERO suites following the protocol of Section~\ref{sec:libero_experiments}: LIBERO seed 7, 10 tasks per suite, 50 trials per task, for 500 episodes per suite. Our heatmap head replaces $\pi_{0.5}$'s flow-matching action expert while preserving the rest of the backbone. Both our heatmap head and the flow-matching baseline are evaluated at 30000 training steps, matching the published $\pi_{0.5}$ training budget.

\paragraph{Finetuning.} The full training recipe is summarized in the LIBERO column of Tab.~\ref{tab:pi_recipe}. We use the JAX-based codebase from the original $\pi_{0.5}$ release. Our heatmap head uses a $64 \times 64 \times 64$ translation grid and a $48 \times 48 \times 48$ rotation grid with Gaussian-blob $\sigma = 0.15$, and decodes by top-$k$ soft-argmax with $k = 10$ at inference (Section~\ref{sec:methodology_inference}). The flow-matching baseline keeps every other hyperparameter fixed.

\begin{table}[htbp]
\centering
\caption{Training recipe for our $\pi_{0.5}$ experiments. The optimizer, learning-rate schedule, warm-up steps, and heatmap-head hyperparameters are shared between the LIBERO simulation and real-world Franka settings; the two settings differ only in global batch size, GPU count, and training-step budget. Our heatmap head replaces $\pi_{0.5}$'s flow-matching action expert; the flow-matching baseline keeps every other hyperparameter fixed.}
\label{tab:pi_recipe}
\begin{tabular}{lcc}
\toprule
 & \textbf{LIBERO} & \textbf{Real-World Franka} \\
\midrule
Framework         & JAX & JAX \\
Optimizer         & AdamW & AdamW \\
LR schedule       & Cosine, $5{\times}10^{-5} \to 3{\times}10^{-5}$ & Cosine, $5{\times}10^{-5} \to 3{\times}10^{-5}$ \\
Warm-up steps     & 1000 & 1000 \\
Global batch size & 256 & 64 \\
GPUs              & 8 $\times$ H200 & 4 $\times$ H200 \\
Training steps    & 30000 & 20000 \\
\midrule
Translation grid  & $64 \times 64 \times 64$ & $64 \times 64 \times 64$ \\
Rotation grid     & $48 \times 48 \times 48$ & $48 \times 48 \times 48$ \\
Gaussian $\sigma$ & 0.15 & 0.15 \\
Decoder           & top-$k$ soft-argmax ($k=10$) & top-$k$ soft-argmax ($k=10$) \\
\bottomrule
\end{tabular}
\end{table}

\section{$\pi_{0.5}$ Real-World Franka Experiments}
\label{sec:appendix_pi_real_world}

\subsection{Experimental Setup}

\paragraph{Tasks and Evaluation.} We evaluate $\pi_{0.5}$ on a separate set of three real-world Franka tasks, distinct from the OpenVLA-OFT tasks of Section~\ref{sec:real_world_franka_setup}:

\begin{enumerate}
    \item \textbf{Block}: \textit{"Place the orange block on top of the gray block"}; 125 demonstrations.
    \item \textbf{Cup}: \textit{"Place the cup on the green coaster"}; 100 demonstrations.
    \item \textbf{Sweep}: \textit{"Sweep the green block into the dustpan"}; 125 demonstrations.
\end{enumerate}

For the data-efficiency setting, we additionally finetune both heads on a 50\% subset of each task's demonstrations (55, 60, and 65 episodes for Block, Cup, and Sweep respectively). For each cell (task $\times$ head $\times$ data scale), we perform 15 rollouts.

\paragraph{Finetuning.} The training recipe is reported in the real-world Franka column of Tab.~\ref{tab:pi_recipe}. The optimizer, learning-rate schedule, warm-up steps, and heatmap-head hyperparameters match the LIBERO setup of Appendix~\ref{sec:appendix_pi_libero}. The differences are a smaller global batch size (64 instead of 256, on 4 H200 GPUs) and a shorter training-step budget (20000 instead of 30000), both reflecting the smaller dataset size of each real-world task.

\subsection{Quantitative Analysis}

Table~\ref{tab:pi_real_world_results} reports per-task and pooled success rates for both heads at the 50\% partial-data setting and at full data. Our heatmap head wins on every task at the partial-data setting (pooled 34/45 vs.\ 29/45 for the flow-matching baseline, $+5$ trials). At full data, the per-task split is balanced: our heatmap head wins on Sweep ($14/15$ vs.\ $12/15$), ties on Cup ($10/15$ each), and trails on Block ($12/15$ vs.\ $14/15$), with the pooled count tied at 36/45. The pattern follows the data-efficiency observation of Section~\ref{sec:exp_main}: our heatmap head's distributional supervision helps most when each demonstration must carry more information, and the gap closes once both heads reach their full-data plateau.

\subsection{Qualitative Analysis}

We conduct a qualitative analysis of our proposed ActionMap across three real-robot manipulation tasks to understand the behavioral characteristics underlying the quantitative gains.

\paragraph{Spatial precision from discrete spatial grounding.} The most consistent behavioral difference between our voxel heatmap head and the flow-matching baseline is spatial precision at contact and placement. In the block-stacking task, the baseline frequently places the orange block slightly off-center, causing it to slide off or topple after release; our method reliably centers the block before descending.  In the cup task, the baseline overshoots the coaster along the approach axis in roughly one-in-five trials, requiring a corrective micro-adjustment that often fails; our method approaches the coaster with a tighter terminal trajectory.  We attribute this to the heatmap head's inductive bias: by predicting the target end-effector position as a probability distribution over a discrete 3-D spatial grid and decoding via soft-argmax, the head is forced to commit to a single, spatially coherent mode rather than averaging over multiple plausible positions in Cartesian space.  Flow matching alone implicitly marginalizes over such modes during the ODE integration, which inflates positional variance at the end of the trajectory where precision matters most.

\paragraph{Motion smoothness from structured $x_{\theta}$ prediction.} Our method produces noticeably smoother end-effector trajectories throughout execution. In the sweeping task that involves a long, continuous arc followed by a sharp directional change into the dustpan, the baseline exhibits oscillatory corrections mid-sweep that occasionally scatter the block outside the dustpan rim; our method executes the arc as a single coherent motion. Similarly, in the block-stacking descent phase, the baseline shows brief z-axis jitter as the block nears the target, whereas our method descends monotonically. The underlying reason is the role of the heatmap head as a structured prior on $x_{\theta}$ within the flow-matching ODE: rather than integrating a velocity field whose $x_{\theta}$ estimate shifts at each denoising step, the heatmap provides a low-variance, spatially regularized anchor.  This constrains the ODE trajectory to a narrower tube in action space, suppressing the high-frequency fluctuations that arise when the velocity field is inconsistent across timesteps.

\paragraph{Residual failure mode: gripper release timing.} Despite the above gains, the dominant remaining error mode across all three tasks is mistimed gripper actuation rather than positioning error.  In block stacking, the gripper occasionally releases before the block is fully seated, resulting in a partial stack that collapses.  In the cup task, early release mid-transfer causes the cup to tip.  In sweeping, the gripper closes too late during the final push, missing the block.  This class of failure is qualitatively distinct from positional errors: the end-effector reaches the correct location, but the binary gripper signal is triggered one or two control steps early or late. We hypothesize two contributing factors.  First, the 7-D action space treats the gripper dimension as a continuous scalar that is thresholded at inference time, losing the fine temporal structure of open/close transitions; a dedicated discrete gripper head could address this.  Second, the 15 Hz control rate means a single-step error corresponds to a 67 ms window—sufficient to cause visible mis-grasps.  Future work should consider the joint prediction of gripper state as a separate classification head conditioned on contact feedback.

\begin{table}[t]
\centering
\caption{Real-world Franka results for $\pi_{0.5}$ on three tasks: Block, Cup, and Sweep. Each cell reports successful trials out of 15 ($n=15$ per cell). Demonstration counts per task at full data and at the 50\% partial-data subset are reported in parentheses next to each task name. Pooled rows sum success counts across the three tasks (out of 45 trials). Our heatmap head wins on every task at the partial-data setting and matches the flow-matching baseline at full data.}
\label{tab:pi_real_world_results}
\begin{tabular}{lcccc}
\toprule
 & \multicolumn{2}{c}{\textbf{50\% Data}} & \multicolumn{2}{c}{\textbf{Full Data}} \\
\cmidrule(lr){2-3} \cmidrule(lr){4-5}
 & $\pi_{0.5}$ & $\pi_{0.5}$ + Ours & $\pi_{0.5}$ & $\pi_{0.5}$ + Ours \\
\midrule
Block (110 / 55) & 12/15 & \textbf{13/15} & \textbf{14/15} & 12/15 \\
Cup (120 / 60)   & 7/15  & \textbf{8/15}  & 10/15 & 10/15 \\
Sweep (130 / 65) & 10/15 & \textbf{13/15} & 12/15 & \textbf{14/15} \\
\midrule
Total            & 29/45 & \textbf{34/45} & 36/45 & 36/45 \\
\bottomrule
\end{tabular}
\end{table}

\section{Qualitative Analysis on OpenVLA-OFT Real-World Experiments}
\label{sec:appendix_qualitative}

This section provides a qualitative view on the OpenVLA-OFT real-world Franka experiments of Section~\ref{sec:exp_main}, complementing the aggregate success rates and grasp-precision statistics reported there. Section~\ref{sec:appendix_qualitative_failure} characterizes the per-trial failure modes of our heatmap head versus OpenVLA-OFT's L1 regression head across the 120 real-robot rollouts (3 tasks $\times$ 2 data scales $\times$ 2 heads $\times$ 10 trials). Section~\ref{sec:appendix_heatmap_viz} visualizes the predicted heatmaps from our head on selected successful rollouts.

\subsection{Failure Modes and Qualitative Improvement}
\label{sec:appendix_qualitative_failure}

\paragraph{Our heatmap head reduces shortcut-style mode collapse rather than eliminating any specific failure mode.} Across the three real-world Franka tasks our heatmap head does not eliminate any particular per-task failure. Instead, it exhibits less \emph{shortcut-style mode collapse} than OpenVLA-OFT's L1 regression head: it more reliably attempts each task stage (descent, alignment, insertion) instead of skipping a stage or freezing in place. We summarize the per-task patterns below.

\paragraph{Pick: same dominant failure, with smaller bias and tighter spread.} Both heads share the dominant Pick failure, in which the gripper closes behind the brick along the $-x$ axis. Our heatmap head reduces the frequency of this localization error and tightens the spatial distribution of grasp positions (Fig.~\ref{fig:pick_pose_error}). One additional failure mode, in which the gripper grasps with a combined $-x$ and $+y$ offset (for example, grasping ``behind and to the left'', or stabbing onto the top of the brick), appears only with the L1 regression head at 50 episodes, and does not appear with our heatmap head at any data scale.

\paragraph{Sweep: disjoint failure modes between heads.} At full data the two heads disagree on which mode dominates. Our heatmap head approaches the brick and makes surface contact in nine of ten trials, oscillating side-to-side on top of the brick; in six of these trials the gripper escapes the contact and completes a successful sweep, while in the remaining three the gripper remains stuck on the brick top until the rollout times out. The L1 regression head, on the same task and brick placements, hovers slightly above the brick without descending to make contact in eight of ten trials, displaying the same behavior in all eight failures. At 50 episodes our heatmap head succeeds on every trial, while the L1 regression head's most prominent failure mode is an explicit shortcut: in five of ten trials the gripper does not descend toward the brick and freezes in place.

\paragraph{Insert: disjoint failure modes at both data scales.} At full data our heatmap head splits its trials among premature insertion outside the slot, premature insertion that the policy then corrects mid-rollout, and clean alignment-then-insertion successes. The L1 regression head, on identical slot placements and at the same checkpoint budget, commits to a fast direct insertion at a $+y$, $-x$ biased location in every trial and never corrects. At 50 episodes our heatmap head still attempts an outside-slot insertion in every trial, matching the L1 regression head's full-data mode, while the L1 regression head, at the same data scale, reaches a $-x$ biased location and stalls without ever initiating an insertion. At the harder data scale our heatmap head therefore fails in the way the L1 regression head fails at full data, while the L1 regression head does not attempt the task at all.

\paragraph{Side observation: training-data scale and our heatmap head act in the same direction.} Reading the same 120 trials along the data-scale axis within each head, the qualitative effect of more training data points in the same direction as the qualitative effect of switching from the L1 regression head to our heatmap head: both push the policy away from terminal-action shortcuts and toward task-plan execution with localization error. The multi-axis offset variants on Pick disappear for the L1 regression head when moving from 50 episodes to full data, and the Insert L1 regression head transitions from never attempting the task at 50 episodes to direct insertion at a wrong location at full data. The advantage that our heatmap head brings on these tasks is therefore co-directional with training-data scaling, rather than orthogonal to it.

\begin{figure}[h]
    \centering
    \includegraphics[width=\linewidth]{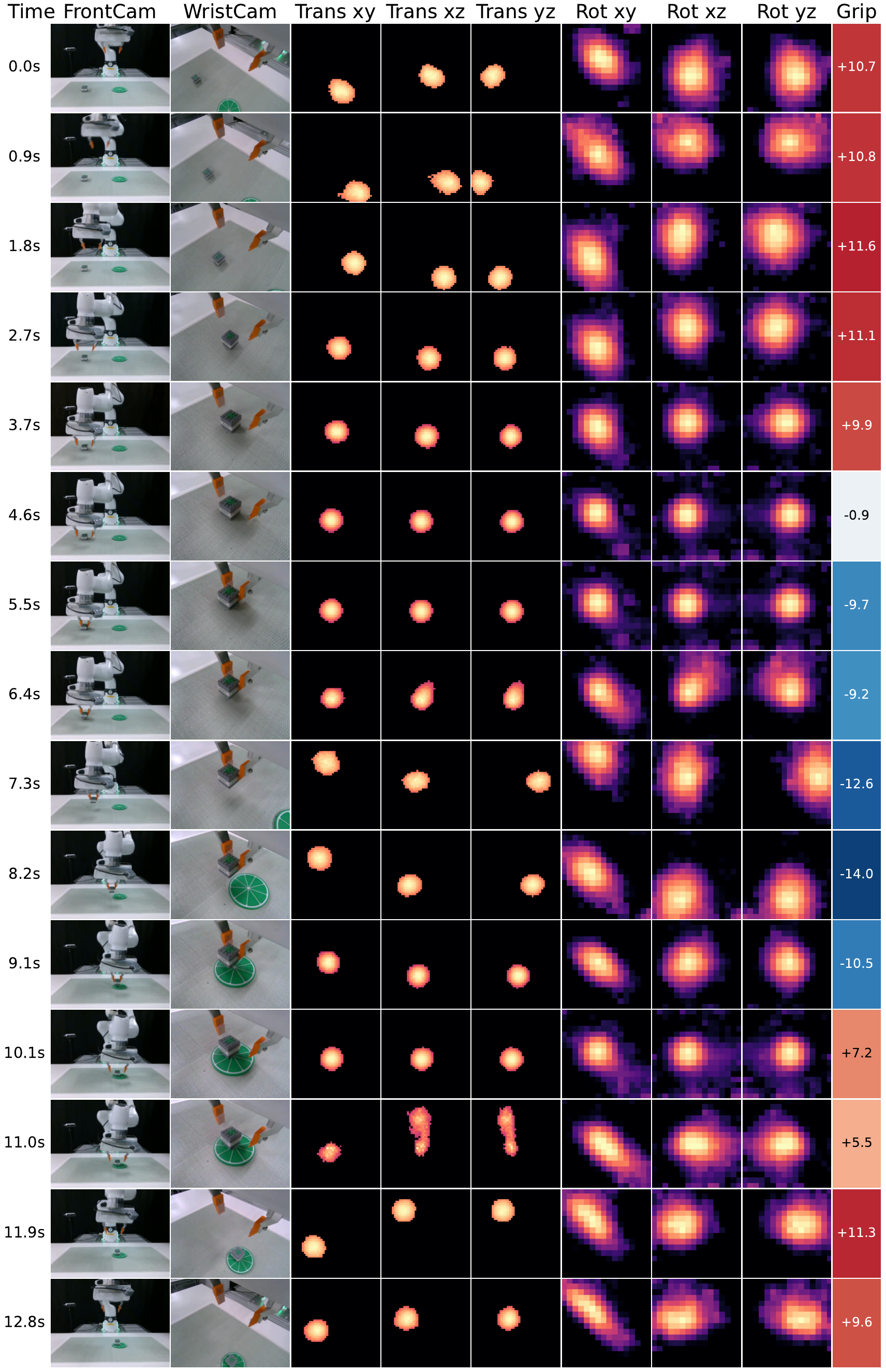}
    \caption{Heatmap visualization on a successful \textbf{Pick} rollout. In each translation and rotation panel, the right axis is the first listed dimension and the top axis is the second listed dimension (e.g., \emph{Trans xy} maps $x$ to the right and $y$ to the top); the center of each panel corresponds to zero displacement, with the robot arm staying at its current pose.}
    \label{fig:heatmap_pick}
\end{figure}

\begin{figure}[h]
    \centering
    \includegraphics[width=\linewidth]{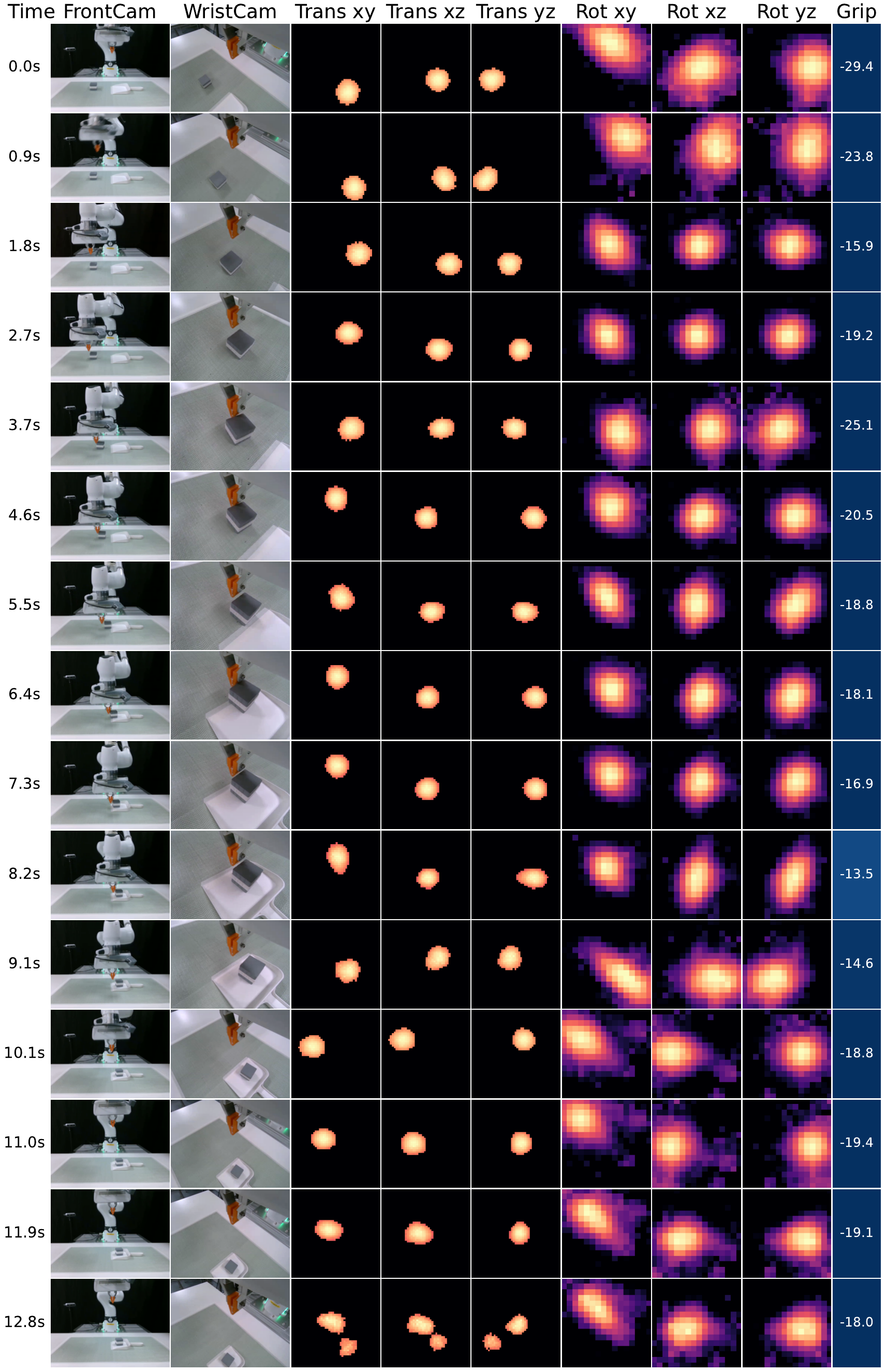}
    \caption{Heatmap visualization on a successful \textbf{Sweep} rollout. In each translation and rotation panel, the right axis is the first listed dimension and the top axis is the second listed dimension (e.g., \emph{Trans xy} maps $x$ to the right and $y$ to the top); the center of each panel corresponds to zero displacement, with the robot arm staying at its current pose.}
    \label{fig:heatmap_sweep}
\end{figure}

\begin{figure}[h]
    \centering
    \includegraphics[width=\linewidth]{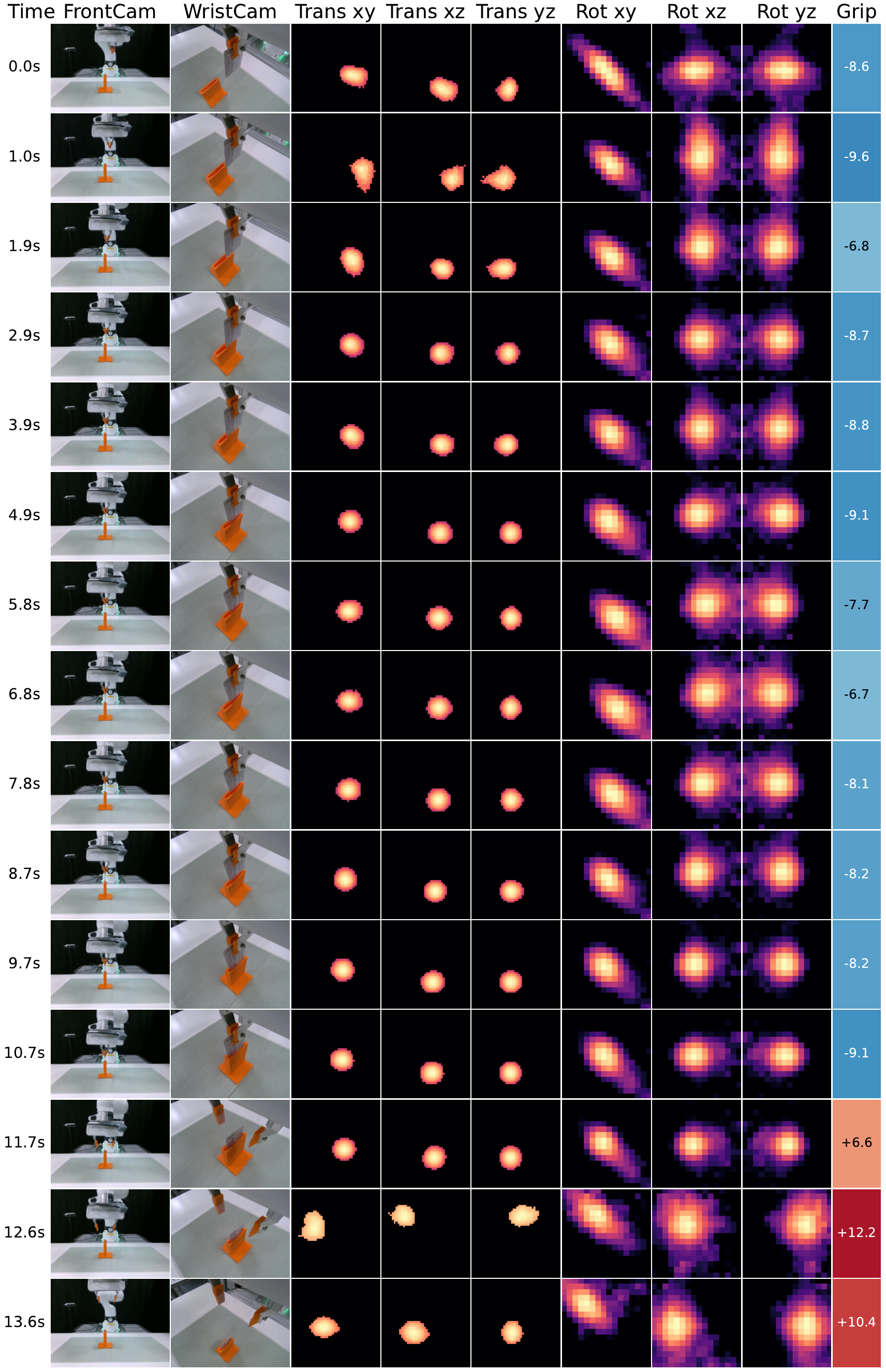}
    \caption{Heatmap visualization on a successful \textbf{Insert} rollout. In each translation and rotation panel, the right axis is the first listed dimension and the top axis is the second listed dimension (e.g., \emph{Trans xy} maps $x$ to the right and $y$ to the top); the center of each panel corresponds to zero displacement, with the robot arm staying at its current pose.}
    \label{fig:heatmap_insert}
\end{figure}

\subsection{Heatmap Visualizations on Successful Rollouts}
\label{sec:appendix_heatmap_viz}

We visualize the predicted heatmaps from successful rollouts of our heatmap head on the three real-world Franka tasks of Section~\ref{sec:real_world_franka_setup}.

\section{Detailed Future Directions}
\label{sec:appendix_future}

We expand here on the four future directions previewed in the conclusion, each addressing a different axis of the design.

\paragraph{Adaptive grid resolutions.} The uniform voxel grid we use in this paper allocates the same resolution to every region of the action space, including regions that are rarely reached or that do not require fine-grained control. A non-uniform grid that allocates higher resolution where training-data density is greatest, or where the task demands precise positioning (such as the contact-rich phases of insertion), would mitigate the cubic cost of finer global resolution. The technical question is what signal should drive the adaptation: training-data density, gradient norms during training, task-phase predictions from the backbone, or per-rollout resampling. A successful adaptive grid would let our heatmap head exploit the headroom we observed at the finest resolution in Section~\ref{sec:ablation_grid_sigma}, without paying the corresponding compute and $\sigma$-tuning cost.

\paragraph{Absolute-coordinate grids.} Our grids are laid out over the delta-action space (per-step velocities), following the convention of recent VLAs \cite{OpenVLA-OFT, Pi-0.5}. An alternative is to voxelize the absolute reachable workspace of the end-effector, as in PerAct \cite{shridhar2022perceiver} and RVT-2 \cite{goyal2024rvt2}. Absolute grids carry richer geometric semantics (a translation bin maps to a fixed Cartesian point) but must also handle larger workspaces, robot-specific reachability bounds, and the loss of chunked-prediction alignment between consecutive action steps. Comparing the two layouts on the same backbone would clarify whether our observed gains are tied to the delta parameterization or carry over to the absolute setting.

\paragraph{Adaptive supervision and alternative target distributions.} Our supervision target is a fixed-$\sigma$ Gaussian blob, and Section~\ref{sec:ablation_grid_sigma} showed that $\sigma$ must be retuned per grid resolution. A natural extension is to make $\sigma$ adaptive: varying with the grid resolution, with the task identity, or with the task phase (e.g., loose $\sigma$ during free-space approach, tight $\sigma$ during contact-rich placement). More broadly, the choice of Gaussian-blob target is itself empirical; alternative distributional targets, such as von Mises distributions over the rotation grid that respect angular topology, may better match the geometry of each branch. Together these would convert the current single-knob $\sigma$ choice into a structured, possibly learned, supervision policy.

\paragraph{Multimodal and temporal sampling.} Our default decoder collapses each per-frame heatmap to a single point via top-$k$ soft argmax. This discards two sources of structure that the heatmap representation makes available: (i) \emph{multimodality} within a single frame, when the demonstrations support multiple valid actions (e.g., a grasp approachable from multiple angles); and (ii) \emph{temporal coherence} across the chunked predictions, where neighboring frames' heatmaps are likely to share peaks. Decoders that sample from the full distribution and enforce temporal smoothness, possibly via a small chain-style filter over the chunk, could improve trajectory smoothness and recover better behavior in ambiguous demonstrations. Section~\ref{sec:ablation_decoding} showed that per-frame decoder choice has small impact, so the headroom here is most likely in the cross-frame and multimodal directions rather than further per-frame variations.

\paragraph{Integration with World Action Models.} Across both autoregressive \cite{wu2023gr1, cheang2024gr2} and diffusion-based \cite{cosmos_policy, dreamzero, fastwam} variants of World Action Models, the action branch still typically performs single-point action prediction, whether regression, autoregressive tokens, or a denoiser. Our voxel heatmap is a drop-in substitute for that branch, leaving the video-prediction branch intact. The combination is attractive in principle: video co-training would supply the world-dynamics signal that VLAs lack, while our heatmap head would supply the spatially structured action representation that current WAM action branches lack. Adjacent directions include scaling training data via human-to-robot video translation and physically grounded video datasets \cite{xhumanoid, h2rgrounder, uenr600k}, extending heatmap action heads to GUI agents \cite{macosworld, incontext_defense}, and generative-model safety tooling relevant to large-scale deployment \cite{ringid, wmadapter, simple_avg_watermarks, idprotector}.



\end{document}